%% file: PaperForReview.tex
\documentclass[10pt,twocolumn,letterpaper]{article}

\usepackage{wacv}              

\usepackage{graphicx}
\usepackage{amsmath}
\usepackage{amssymb}
\usepackage{booktabs}

\usepackage{url}
\usepackage{epsfig}
\usepackage{graphicx}
\usepackage{pifont}
\newcommand{\cmark}{\ding{51}}%
\newcommand{\xmark}{\ding{55}}%
\usepackage{color}
\usepackage{multirow}
\usepackage{xcolor,colortbl}
\usepackage{multicol}
\usepackage{wrapfig,lipsum,booktabs}
\usepackage{xspace}
\usepackage{amsmath}
\usepackage{enumitem}

\usepackage[pagebackref,breaklinks,colorlinks]{hyperref}

\usepackage[capitalize]{cleveref}
\crefname{section}{Sec.}{Secs.}
\Crefname{section}{Section}{Sections}
\Crefname{table}{Table}{Tables}
\crefname{table}{Tab.}{Tabs.}

\def\wacvPaperID{1875} 
\def\confName{WACV}
\def\confYear{2025}

\newcommand{\methodname}{NeuManifold}

\newcommand{\neumanifold}[0]{NeuManifold\xspace}

\newcommand{\boldstart}[1]{\noindent\textbf{#1}}
\newcommand{\boldstartspace}[1]{\vspace{0.1in}\noindent\textbf{#1}}

\newcommand{\xyred}{\cellcolor{red!25}}
\newcommand{\xygreen}{\cellcolor{green!25}}
\newcommand{\xyyellow}{\cellcolor{yellow!25}}

\begin{document}

\title{NeuManifold: Neural Watertight Manifold Reconstruction with Efficient and High-Quality Rendering Support}

\author{Xinyue Wei$^{1}$, Fanbo Xiang$^{1}$, Sai Bi$^{2}$, Anpei Chen$^{3,4}$, Kalyan Sunkavalli$^{2}$\\ Zexiang Xu$^{2*}$, Hao Su$^{1*}$ \\ \\
$^{1}$University of California San Diego, $^{2}$Adobe Research,
$^{3}$ETH Zürich,
$^{4}$University of Tübingen
}

\maketitle

\footnotetext[1]{Research partially done when X. Wei was an intern at Adobe Research}
\footnotetext[2]{* Equal advisory.}

\begin{abstract}
   While existing volumetric rendering approaches provide photorealistic results, extracting high-quality meshes from optimized neural field representations is challenging. Conversely, existing differentiable rasterization-based methods are typically sensitive to initialization and suffer from poor mesh rendering quality. In this paper, we introduce NeuManifold, a novel method for reconstructing watertight manifold meshes with high-quality textures from multi-view input images. NeuManifold overcomes the limitations of existing approaches by first learning a neural volumetric field and then refining it through differentiable mesh extraction and surface rendering. To eliminate artifacts and preserve mesh properties during iso-surface extraction, we introduce a novel differentiable marching cubes method. Instead of traditional textures, we use neural textures to enhance rendering quality. To integrate with modern graphics rendering pipelines, we also provide customized GLSL shader support for neural textures. Extensive experiments demonstrate that NeuManifold outperforms existing mesh-based reconstruction methods in both mesh quality and rendering metrics, achieving comparable or superior rendering quality to prior volume-rendering-based methods. The generated results enable real-time, high-quality rendering and seamlessly support numerous graphics pipelines and applications requiring high-quality meshes, such as 3D printing and physical simulation. \url{https://sarahweiii.github.io/neumanifold/}.
\end{abstract}


\begin{figure*}
  \centering
  \includegraphics[width=0.98\textwidth]{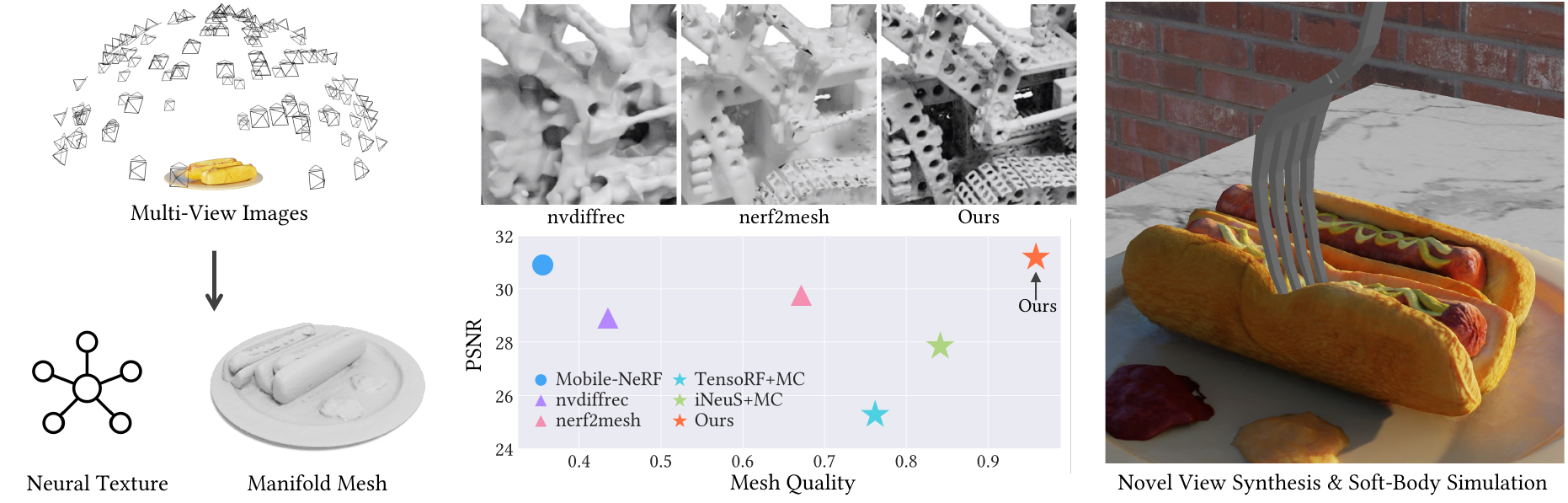}
  \vspace{-0.8em}
  \caption{\textbf{Left}: \neumanifold takes multi-view posed images and generates watertight manifold meshes with neural textures. \textbf{Middle}: \neumanifold outperforms many mesh reconstruction methods in geometry while maintaining excellent rendering quality. The circle, triangle, and star represent methods without a guarantee, methods generating watertight meshes, and methods producing watertight manifold meshes, respectively. \textbf{Right}: The generated manifold meshes with neural textures support downstream applications like high-quality novel-view synthesis and soft-body simulation.}
  \label{fig:teaser}
  \vspace{-1.2em}
\end{figure*}

\section{Introduction}
\input{sections/intro.tex}

\section{Related Work}
\input{sections/related.tex}


\section{Method}
\input{sections/method.tex}

\section{Experiments}
\input{sections/exp.tex}

\section{Applications}

\input{sections/app.tex}

\section{Conclusion}
\input{sections/conclusion.tex}

{\small
\bibliographystyle{ieee_fullname}
\bibliography{egbib}
}

\clearpage
\appendix
\input{sections/supp.tex}

\end{document}

%% file: sections/intro.tex
Recent advancements in neural field representations~\cite{mildenhall2021nerf,muller2022instant,chen2023factor} have enabled scene reconstructions with photorealistic rendering quality. However, these methods typically rely on implicit volumetric representations, resulting in slower processing speeds and limited compatibility with standard graphics pipelines, such as 3D printing, geometry editing, and physical simulation.

For many such applications, meshes—especially those that are manifold and watertight—are the preferred option. Meshes can be efficiently rendered using standard 3D rendering engines. Additionally, the watertight manifold property is often advantageous in various geometry processing algorithms, such as mesh Boolean operations, approximate convex decomposition~\cite{wei2022coacd}, mesh tetrahedralization~\cite{hang2015tetgen,hu2018tetrahedral,hu2020fast}, and volumetric point sampling for initializing particle simulations~\cite{stomakhin2013material,bender2017survey,solenthaler2009predictive}.

Although mesh reconstruction has been extensively studied in prior arts~\cite{schoenberger2016mvs,snavely2006photo,furukawa2010accurate}, reconstructing a high-quality manifold mesh with realistic rendering remains a significant challenge. In response, recent advancements in inverse graphics through differentiable surface rendering have shown considerable promise, such as nvdiffrec~\cite{munkberg2021nvdiffrec} and nerf2mesh~\cite{tang2022nerf2mesh}. 
However, the rendering quality of these methods still falls short of neural-field-based approaches, and their meshes, optimized primarily for rendering applications, often result in non-manifold models with self-intersections. These are unsuitable for 3D printing, physical simulation, and other geometry processing applications.

To bridge this gap, we introduce \methodname{}, a novel approach for reconstructing high-quality, \emph{watertight manifold} meshes of 3D objects with neural textures from posed multi-view images. \methodname{} combines advanced volumetric neural field rendering with differentiable rasterization mesh reconstruction techniques, offering mutually complementary benefits. Specifically, NeuManifold begins by learning a neural volumetric field, facilitating flexible geometry topology and high visual quality. Extracting high-quality mesh from the optimized implicit field while retaining the original high visual quality achieved through volume rendering is non-trivial. Therefore, we use the learned results as a solid initial point and further refine them with differentiable mesh extraction and surface rendering. Existing rasterization-based approaches, such as nvdiffrec \cite{munkberg2021nvdiffrec}, which optimize the final mesh representation from scratch, are sensitive to geometry initialization and often get trapped in local minima, especially when reconstructing high-resolution meshes (see Fig.~\ref{fig:mesh_comparison}). In contrast, \methodname{} benefits from high-quality initializations, significantly improving the final mesh reconstruction quality.

To refine the geometry networks using differentiable rasterization, an important component is needed: differentiable iso-surface extraction. The most commonly used algorithm for this purpose has been the differentiable tetrahedra-based algorithm DMTet~\cite{shen2021deep}. However, we have observed that the non-linear characteristics of the density field can lead to undesirable artifacts when using DMTet. In contrast, voxel-based algorithms, such as Marching Cubes, can significantly reduce these artifacts, as illustrated in Fig.~\ref{fig:diffmc}. For end-to-end training of the networks, differentiable voxel-based iso-surface algorithms are essential. However, existing methods such as FlexiCubes~\cite{shen2023flexible} do not adequately preserve manifold properties. To address this issue, we introduce Differentiable Marching Cubes (DiffMC), which effectively eliminates these artifacts and results in significantly smoother surfaces. 
Moreover, we provide an efficient DiffMC implementation with CUDA, which is substantially more efficient in terms of both GPU memory and speed compared to previous works (see Tab.~\ref{tab:diffmc_speed}).

Instead of using the traditional BRDF textures utilized in most inverse rendering methods~\cite{munkberg2021nvdiffrec,luan2021unified}, we opt to use neural textures for appearance modeling of the generated mesh, which enhances the visual quality to the greatest extent possible. To seamlessly integrate into modern graphics rendering pipelines, we also provide customized GLSL shader support for the generated neural textures.


Extensive experiments on both synthetic and real data have demonstrated that \methodname{} surpasses existing mesh-based reconstruction methods in terms of both mesh quality metrics and rendering metrics. It achieves comparable rendering quality with prior volume-rendering-based methods and significantly higher quality than mesh-based methods. For instance, it attains a 2.29 dB higher PSNR than nvdiffrec~\cite{munkberg2021nvdiffrec}, along with superior mesh properties, as illustrated in Fig.~\ref{fig:teaser}. Furthermore, the results generated by \methodname{} enable real-time, high-quality rendering and support numerous graphics applications with stringent mesh requirements, such as 3D printing and physical simulation. In summary, our work offers the following key contributions:



\begin{itemize}[leftmargin=*]
\item We propose \methodname{}, which excels in producing high-quality, watertight manifold meshes with neural textures. The generated results can be seamlessly integrated with existing graphic pipelines and support real-time, realistic rendering, and numerous downstream applications.
\item We introduce the first complete model of Differentiable Marching Cubes (DiffMC). 
Our DiffMC delivers smooth surfaces from density fields and effectively preserves both manifoldness and watertightness.
Notably, our CUDA-based implementation leads to a significant speedup ($\sim$20x) and a substantial reduction in memory usage ($\sim$1/5) compared to DMTet and FlexiCubes. 
 
\item We enhance rendering quality using neural textures and implement them in the GLSL shader, allowing seamless integration of our mesh-based representation into modern rendering engines.
\item Extensive experiments demonstrate that our method achieves high-quality rendering and mesh reconstruction simultaneously, significantly outperforming existing mesh-based methods.

\end{itemize}





%% file: sections/related.tex
\subsection{Neural field representations.}
Neural rendering methods have demonstrated photorealistic scene reconstruction and rendering quality. 
In particular, NeRF \cite{mildenhall2021nerf} introduced the neural radiance field representation and achieved remarkable visual quality with volume rendering techniques. 
Various neural field representations have been proposed for better efficiency and quality, including MipNeRF \cite{barron2021mip} and RefNeRF \cite{verbin2022ref} that are based on coordinate-based MLPs, TensoRF \cite{chen2022tensorf} and DiF\cite{chen2023dictionary} that leverage tensor factorization, iNGP \cite{muller2022instant} that introduces multi-scale hashing, Plenoxels\cite{fridovich2022plenoxels} and DVGO\cite{sun2022direct} that are voxel-based, and Point-NeRF \cite{xu2022point} that is based on neural point clouds. 

However, most neural field representations represent 3D geometry as a volume-density field, which is hard to edit for 3D applications.
While several methods have enabled appearance editing or shape deformation \cite{xiang2021neutex, zhang2022arf, yuan2022nerf, wu2022palettenerf,kuang2022palettenerf,neumesh}, it is still highly challenging to apply them directly in modern 3D engines.
Recent methods have proposed replacing density fields with volume SDFs to achieve better surface reconstruction with volume rendering \cite{wang2021neus,wang2022neus2,yariv2021volume,Oechsle2021ICCV}.
However, Density- or SDF-based models struggle to be exported as meshes without notable loss in \emph{rendering} quality. MobileNeRF \cite{chen2022mobilenerf} turns the neural field into a triangle soup for real-time rendering but fails to capture accurate scene geometry for further applications. Our work provides a versatile solution for converting volumetric neural fields into high-quality manifold meshes, facilitating both high-quality rendering and a range of 3D applications such as physical simulation.


\subsection{Mesh reconstruction and rendering.}

Polygonal meshes are a staple in modern 3D engines, widely employed for modeling, simulation, and rendering. Previous research has extensively explored mesh reconstruction from multi-view captured images through photogrammetry systems\cite{pollefeys2002images,snavely2006photo,schoenberger2016mvs} like structure from motion\cite{schoenberger2016sfm,tang2018ba,vijayanarasimhan2017sfm}, multi-view stereo\cite{furukawa2010accurate,kutulakos2000theory,schoenberger2016mvs,yao2018mvsnet,cheng2020deep}, and surface extraction techniques\cite{lorensen1987marching,kazhdan2006poisson}. However, achieving photorealistic rendering with classical photogrammetry pipelines remains a formidable challenge.

On the other hand, inverse rendering aims to fully disentangle intrinsic scene properties from captured images 
\cite{goldman2009shape,hernandez2008multiview,zhang2021nerfactor,bi2020deepref, bi2020deep,bi2020neural,zhang2021physg, li2018learning, zhang2022iron,InvRenderer}. 
Recent methods, such as nvdiffrec \cite{munkberg2021nvdiffrec}, nerf2mesh\cite{tang2022nerf2mesh}, BakedSDF \cite{yariv2023bakedsdf}, achieve high-quality reconstruction and fast rendering speed.
Nevertheless, these methods often introduce self-intersections or an excessive number of triangles in the mesh reconstruction, which are undesired for simulation and geometry processing tasks.

\subsection{Differentiable Iso-Surface Extraction}

In recent years, there has been increasing interest in differentiable iso-surface extraction algorithms due to their crucial role in bridging implicit fields with explicit mesh representations, enabling end-to-end training using differentiable rasterization. Deep Marching Cubes~\cite{liao2018deep} proposed an alternative to traditional marching cubes for extracting meshes from network-predicted occupancy. NMC\cite{chen2021neural} and NDC~\cite{chen2022neural} train neural networks on large-scale datasets, achieving better preservation of shape features. These learning-based methods often sacrifice mesh manifold properties.  MeshSDF~\cite{remelli2020meshsdf} approaches differentiable processing through the gradient of SDF, and ~\cite{mehta2022level} expanded this using level-set theory, but they struggle to maintain sharp features. DMTet~\cite{shen2021deep} and FlexiCubes~\cite{shen2023flexible} explicitly make Marching Tetraheda~\cite{doi1991efficient} and Dual Marching Cubes~\cite{nielson2004dual} differentiable, respectively, and introduce learnable modules to enhance performance. However, FlexiCubes can introduce self-intersections, and DMTet suffers from artifacts when applied to density fields. Our method, DiffMC, addresses these challenges; it can accurately fit the original shape while preserving all mesh properties.

%% file: sections/method.tex
\begin{figure*}[ht]
\begin{center}
   \includegraphics[width=\linewidth]{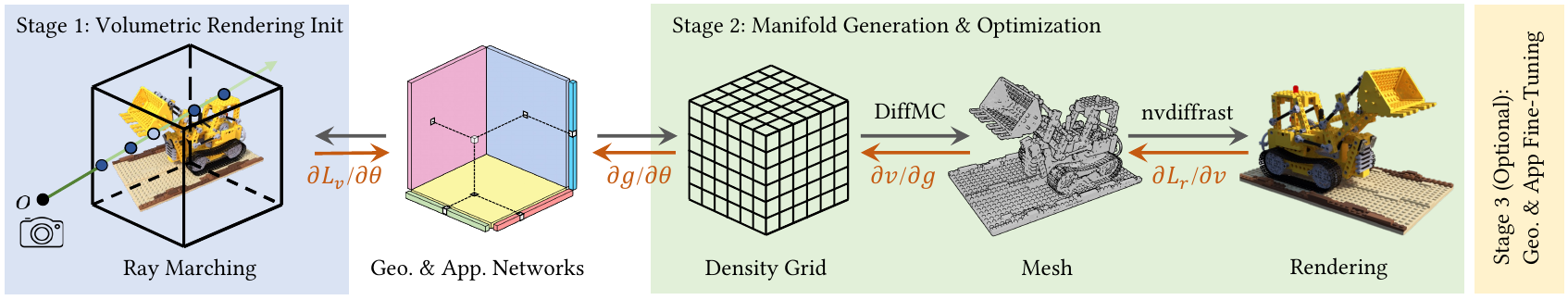}
\end{center}
\vspace{-1.8em}
  \caption{\textbf{Overall training pipeline of \neumanifold.} In Stage 1, volumetric rendering pipelines are used to initialize geometry and appearance networks. In Stage 2, the initialized geometry and appearance networks are further trained in differentiable rasterization with the help of DiffMC, generating watertight manifold mesh. In the optional Stage 3, the mesh vertices are moved to further improve rendering quality. }
\label{fig:pipeline}
\vspace{-1em}
\end{figure*}

We present NeuManifold, a novel 3D reconstruction pipeline that reconstructs watertight manifold meshes with high-quality textures from captured multi-view images.

Existing neural-field-based methods~\cite{chen2022tensorf,muller2022instant} can achieve high rendering quality and effectively support flexible topology changes using neural density field representations. However, extracting high-quality meshes from these learned density fields is challenging due to their lack of a precise surface definition. Directly using iso-surface extraction methods like marching cubes (MC) on the density field necessitates the addition of a margin to artificially create a pseudo-surface. This may result in the extracted meshes being larger than the actual surface and significantly reducing rendering quality, as discussed in Sec.~\ref{sec:sota}. Conversely, existing mesh-based methods~\cite{munkberg2021nvdiffrec} facilitate accurate surface extraction, but the mesh representation imposes strict constraints during the optimization process. This can lead to the optimization getting trapped in local minima, particularly when reconstructing high-resolution meshes, as detailed in Sec.~\ref{sec:stage1}. Such limitation hinders the reconstruction of high-quality meshes with fine details. Fortunately, we discovered that with proper initialization, mesh-based methods can converge to a more favorable state, enabling the reconstruction of high-quality meshes. Therefore, we propose to integrate the advantages of both approaches.



As depicted in Fig.~\ref{fig:pipeline}, the proposed method comprises two primary stages. In the first stage, we optimize a neural field representation using volume rendering to provide a solid initialization. During the subsequent mesh optimization phase, we further optimize the \emph{topology}, \emph{geometry}, and \emph{appearance}. This is achieved by extracting the iso-surface with our proposed differentiable marching cubes and rendering with differentiable rasterization. Optionally, we further fine-tune the \emph{geometry} and \emph{appearance} by directly adjusting mesh vertices and the appearance network, which enhances the rendering quality. Since neural textures, rather than traditional textures, are used, we also provide rendering support for the generated results using GLSL shaders, which enables cross-platform, real-time rendering. We elaborate on the details of each stage from Sec.~\ref{sec:stageone} to Sec.~\ref{sec:stagethree} and illustrate how we deploy the generated results in modern rendering engines in Sec.~\ref{sec:deploy}.


\subsection{Stage 1: Initialization by Volume Rendering}\label{sec:stageone}
In the first stage, we train the networks through differentiable volume rendering to establish a strong initialization for the subsequent differentiable rasterization-based optimization phase. 

We represent a 3D scene with a geometry network $G$ and an appearance network $A$. Given a3D location $x$, the geometry network outputs its corresponding volume density $\sigma$, and the appearance network regresses a view-dependent color $c$ at the location: 
\begin{equation}
\vspace{-0.5em}
    \sigma_{x}, c_x = G(x), A(x,d)
\end{equation}
where $d$ is the viewing direction.
Our approach supports any common neural field representations for the geometry and appearance networks. In this work, we choose the state-of-the-art neural field representation TensoRF~\cite{chen2022tensorf}.


As in NeRF, we render pixel colors $C$ using the volume density and view-dependent colors from geometry and appearance models: 
\begin{equation}
\centering
\vspace{-0.5em}
        C =  \sum^N_{i=1} T_i (1-\exp (-\sigma_i \delta_i)) c_i, \quad 
        T_i = \exp (-\sum_{j=1}^{i-1} \sigma_j \delta_j).
    \label{eq:raymarching}
\end{equation}
where $T$ is the volume transmittance and $\delta$ is the ray marching step size. This differentiable rendering process allows us to optimize our networks with a rendering loss.

\subsection{Stage 2: Manifold Generation \& Optimization}
\label{sec:stagetwo}


\begin{figure}[ht]
\begin{center}
   \includegraphics[width=\linewidth]{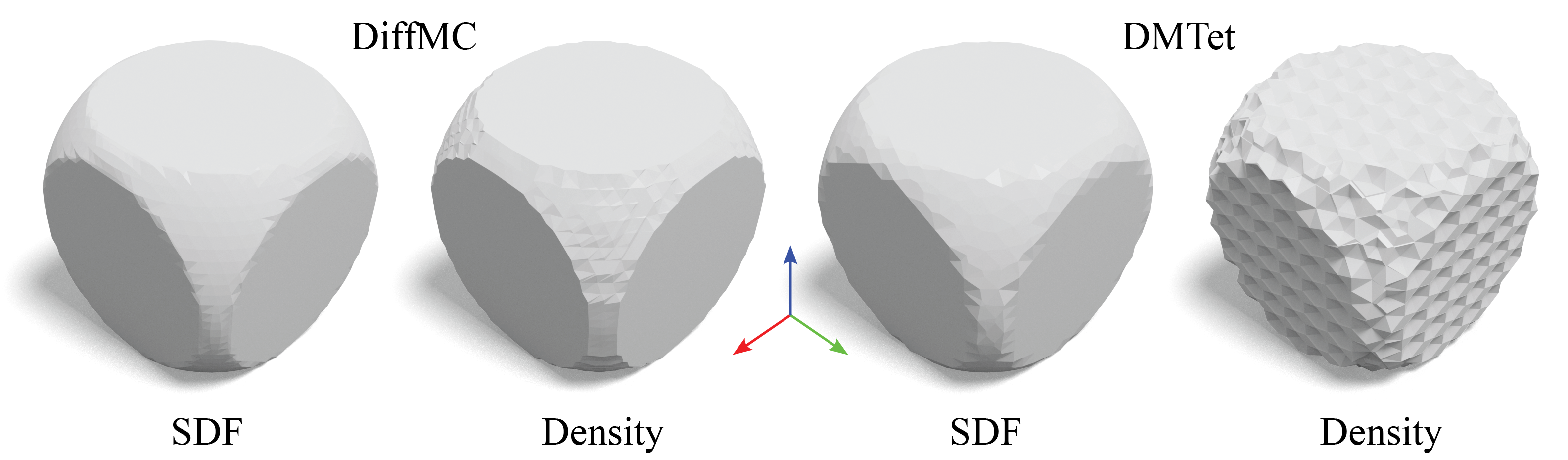}
\end{center}
\vspace{-1.2em}
  \caption{Comparison of the voxel-based method DiffMC and the tetrahedra-based method DMTet on SDF and density fields. DMTet significantly struggles with the non-linear nature of density fields, resulting in the generation of severe artifacts. }
\label{fig:diffmc}
\vspace{-0.2em}
\end{figure}

In the second stage, we utilize the network weights pre-trained in the first stage as initializations. We then employ differentiable rasterization to optimize the object topology, geometry, and appearance simultaneously. 

In this stage, the implementation of a differentiable iso-surface method was necessary. Initially, we integrated DMTet~\cite{shen2021deep} into our architecture due to its effective preservation of watertight manifold properties. However, we observed that DMTet produces severe artifacts when extracting meshes from density fields. As depicted in Fig.\ref{fig:diffmc}, meshes extracted from density fields using DMTet exhibit deformations characterized by peaks and valleys, an issue not present with SDF. We attribute this to the non-linear characteristics of density fields. The mesh vertex generation of DMTet relies on linear interpolations between pairs of tetrahedra vertices. Consequently, the traditional method of linear interpolation proves inadequate for handling non-linear fields, leading to surface artifacts, particularly on surfaces not aligned with tetrahedral divisions. Considering that most real-world objects are typically axis-aligned, adopting an axis-aligned spatial division can significantly mitigate these artifacts. This explains why voxel-based methods are effective on density fields. For further elucidation, a 2D example is provided in Appendix Fig. 1.

Therefore, we introduce the Differentiable Marching Cubes method (DiffMC), which operates on an axis-aligned grid. DiffMC not only extracts meshes using the conventional Marching Cubes~\cite{lorensen1998marching} but also provides gradients for the mesh vertices relative to the grid. Thanks to the advantageous properties of Marching Cubes~\cite{lorensen1998marching}, the generated mesh effectively preserves watertightness, manifoldness, and intersection-free characteristics.
As illustrated in Fig.~\ref{fig:diffmc}, DiffMC is less susceptible to the non-linear nature of the data and is capable of producing significantly smoother surfaces, even for geometries not aligned with the grid axis.

Furthermore, we have implemented the algorithm using CUDA for optimal performance, enabling efficient GPU operation. To the best of our knowledge, we are the first to implement a complete differentiable marching cubes algorithm, achieving speeds approximately 20 times faster than DMTet.

The resulting mesh from DiffMC is fed into nvdiffrast~\cite{Laine2020diffrast} to render 2D images. We then use the rendering loss to update the geometry and appearance networks. With a solid initialization from networks pretrained in volume rendering and the proposed voxel-based iso-surface extraction algorithm, we achieve watertight manifold meshes that are more accurate than both volumetric rendering and mesh rendering alone, offering superior visual quality.

\begin{figure*}[ht]
\begin{center}
   \includegraphics[width=\linewidth]{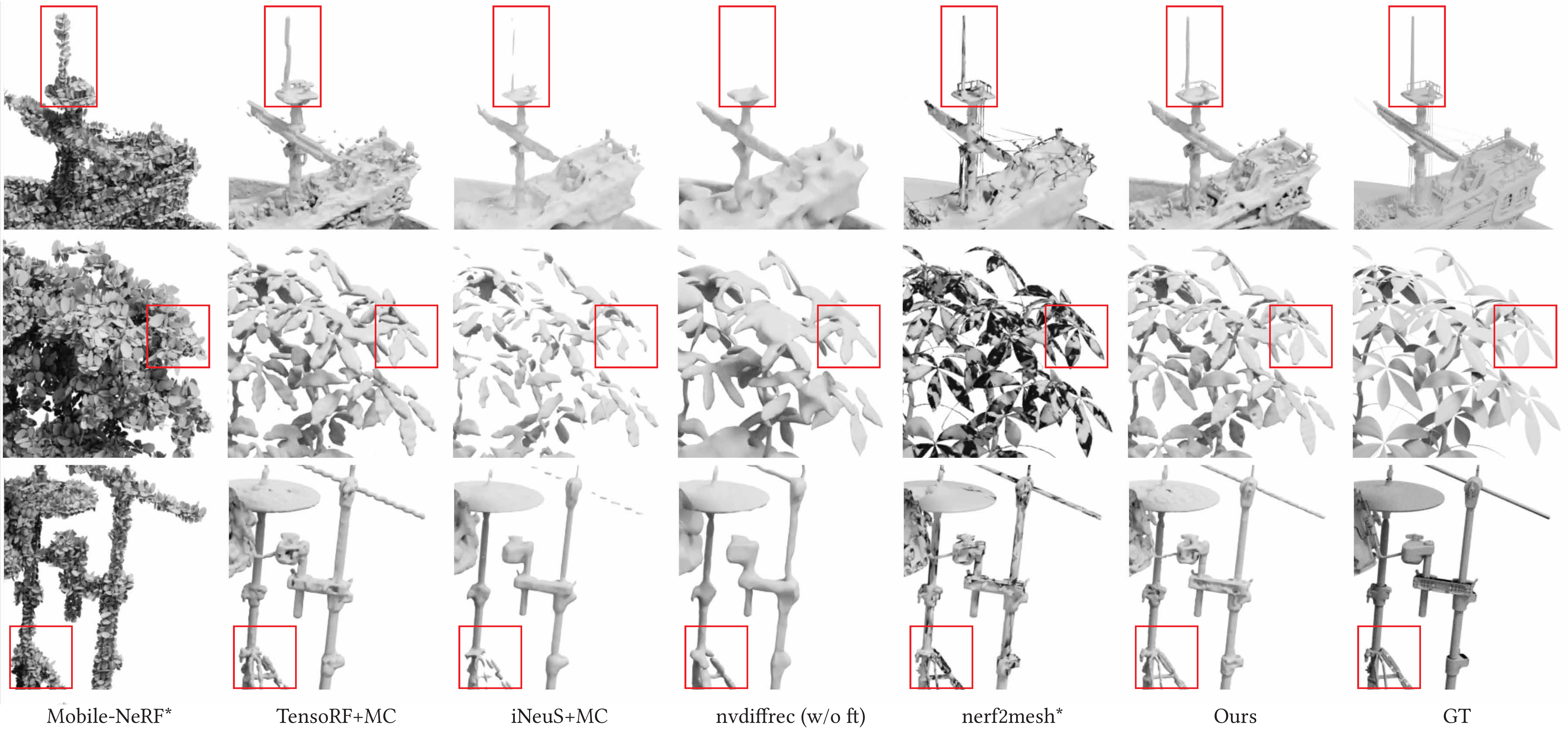}
\end{center}
\vspace{-1.5em}
  \caption{\textbf{Visual comparison of mesh quality across different methods.} MobileNeRF produces a 'triangle soup' that preserves only the rough shape. Nvdiffrec yields a coarse mesh and fails in some regions. TensoRF generates meshes with high-frequency noise, while NeuS tends to be over-smoothed, losing detail. Our method combines the strengths of these approaches, yielding results that are comparable to, or even better than, the non-manifold meshes produced by nerf2mesh. (Methods producing non-manifold meshes are denoted by *, and wrongly oriented faces are marked in black color.)}
\label{fig:mesh_comparison}
  \vspace{-0.8em}
\end{figure*}

\begin{figure*}[ht]
\includegraphics[width=\linewidth]{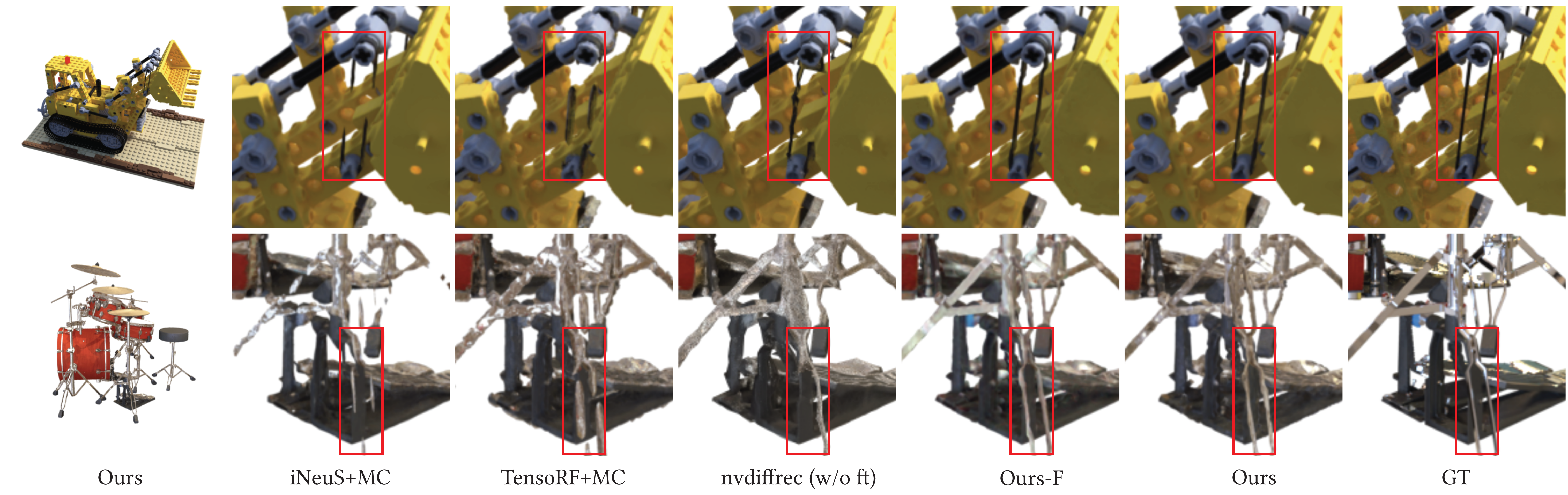}
\vspace{-1.8em}
\caption{\textbf{Rendering quality comparison between our and existing mesh rendering methods.} Our methods can well preserve thin structures as well as achieve high rendering quality.}
\label{fig:sota}
\vspace{-1em}
\end{figure*}

\subsection{Stage 3: Geometry and Appearance Fine-tune}\label{sec:stagethree}

The mesh generated by DiffMC is guaranteed to be a watertight manifold, meeting the stringent requirements of common geometry processing algorithms. However, the degree of freedom in DiffMC is limited. To enhance rendering quality, we can directly adjust the positions of mesh vertices. Specifically, we fine-tune only the positions of the mesh vertices and the appearance network to minimize the rendering loss. This approach preserves the original robust edge connections, thereby maintaining watertightness. However, it may occasionally introduce self-intersections.



\subsection{Rendering Deployment}\label{sec:deploy}
\boldstartspace{GLSL shaders.}
Our pipeline generates a triangle mesh accompanied by an appearance network. We have implemented the neural networks as GLSL shaders, allowing the results produced by our method to be seamlessly integrated into modern rendering engines. The appearance network is composed of TensoRF and MLPs. We treat the TensoRF weights as three 3D textures and three 1D textures with linear filtering, and the MLP weights as 2D textures. After rasterizing the triangles in the vertex shader, we compute TensoRF and the MLPs in the fragment shader using model-space coordinates and viewing directions.


\boldstartspace{Anti-Aliasing. }
Aliasing is a prevalent issue in rasterization pipelines, often arising from the undersampling of high-frequency features, such as mesh edges and detailed textures. Unlike volumetric rendering, which uses semi-transparent volumes to reduce aliasing, mesh-based pipelines are more affected by this issue. We address this by using Supersample Anti-Aliasing (SSAA), which renders at high resolution and downsamples to the target resolution, offering excellent visual quality but at a higher computational cost. An efficient alternative is Multisample Anti-Aliasing (MSAA), which reduces aliasing by averaging multiple samples per pixel, a feature supported by modern GPUs. 


\input{tables/sota_render.tex}

%% file: tables/sota_render.tex
\begin{table*}[t]
\small
\begin{center}
    \caption{\textbf{Average results on NeRF-Synthetic dataset.} The results of NeRF and MobileNeRF are taken from their papers, and the other results for mesh rendering are tested on our machine using the official implementation. The \textit{geometry property} is grouped by color.}
    \label{tab:sota}
\vspace{-1em}
\begin{tabular}{l|cc|cccc|ccc}
\toprule
Method & Geometry & Render & Mesh & Watertight & Manifold & VSA(0.05)$\uparrow$& PSNR$\uparrow$ & SSIM$\uparrow$ & LPIPS$\downarrow$\\
\hline
NeRF~\cite{mildenhall2021nerf} & Volume & RayMarch & \xmark & - & -  & - & 31.00 & 0.947 & 0.081 \\
TensoRF~\cite{chen2022tensorf} & Volume& RayMarch & \xmark & - & - & - & 33.20 & 0.963 & 0.050 \\
iNeuS~\cite{instant-nsr-pl} & Volume& RayMarch & \xmark & - & - & - & 30.74 & 0.951 & 0.064 \\
\midrule
MobileNeRF~\cite{chen2022mobilenerf} & TriSoup & Rasterize & \xyred\cmark & \xyred\xmark & \xyred\xmark & 0.559 & 30.90 & 0.947 & 0.062 \\
nvdiffrec~\cite{munkberg2021nvdiffrec} & Mesh & Rasterize & \xyyellow \cmark & \xyyellow \cmark &  \xyyellow \xmark & 0.633 & 28.90 & 0.938 & 0.073 \\
nerf2mesh~\cite{tang2022nerf2mesh} & Mesh & Rasterize & \xyyellow \cmark & \xyyellow \cmark &  \xyyellow \xmark & 0.787 & 29.76 & 0.940 & 0.072 \\
TensoRF+MC & Mesh & Rasterize & \xygreen \cmark & \xygreen \cmark & \xygreen \cmark & 0.827 & 25.28 & 0.886 & 0.115 \\
iNeuS+MC & Mesh & Rasterize & \xygreen \cmark & \xygreen \cmark & \xygreen \cmark & 0.856 & 27.85 & 0.935 & 0.074 \\
nvdiffrec (w/o ft) & Mesh & Rasterize & \xygreen \cmark & \xygreen \cmark & \xygreen \cmark & 0.635 & 27.65 & 0.933 & 0.084 \\
\midrule
Ours-F (w/ ft) & Mesh & Rasterize & \xyyellow \cmark & \xyyellow \cmark &  \xyyellow \xmark & 0.872 & 30.94 & 0.952 & 0.061 \\
Ours (w/ ft) & Mesh & Rasterize & \xyyellow \cmark & \xyyellow \cmark &  \xyyellow \xmark & \textbf{0.899} & \textbf{31.65} & \textbf{0.956} & \textbf{0.056} \\
Ours-F & Mesh & Rasterize & \xygreen \cmark & \xygreen \cmark & \xygreen \cmark & 0.860 & 30.47 & 0.949 & 0.065 \\
Ours & Mesh & Rasterize & \xygreen \cmark & \xygreen \cmark & \xygreen \cmark & \underline{0.890} & \underline{31.19} & \underline{0.954} & \underline{0.059} \\
\bottomrule
\end{tabular}
\end{center}

\vspace{-2em}
\end{table*}

%% file: sections/exp.tex
We present extensive experiments demonstrating that our method achieves high-quality mesh reconstruction (Sec. \ref{sec:mesh_recon}), photorealistic rendering quality (Sec. \ref{sec:sota}), and real-time rendering (Sec. \ref{sec:real_time}). Additionally, we have conducted numerous ablation studies to illustrate the significance of each module in our pipeline (Sec. \ref{sec:stage1}).

We compare our method with state-of-the-art mesh-based methods.
Additionally, we adapt two volume-based methods, TensoRF~\cite{chen2022tensorf} and Instant NeuS (iNeuS) \cite{instant-nsr-pl}, for surface rendering to provide further comparison. Specifically, we use Marching Cubes (MC) to extract meshes and, during rasterization, query the colors of the surface points from their appearance networks. Thus, we refer to the adapted methods as TensoRF+MC and iNeuS+MC, respectively. We offer two versions of our method differing in their appearance networks: one prioritizes higher quality (Ours) and the other faster rendering speed (Ours-F). The parameters of their appearance networks and the rendering speeds are detailed in Sec.~\ref{sec:real_time}. 


\subsection{Comparison on Mesh Reconstruction}\label{sec:mesh_recon}
We have observed that traditional mesh-distance metrics, such as the Chamfer distance, are not well-suited for comparing mesh quality, as they often disproportionately reflect the performance in regions not seen during training. To this end, we propose to use the visible surface agreement (VSA) metric, modified from the visible surface discrepancy proposed by~\cite{hodavn2020bop}:
\[e_{\text{VSA}} = \substack{\mbox{avg}\\{p\in V\cup \hat V}}
\begin{cases}
1 & \mbox{if } p \in V\cap \hat V \wedge |D(p)-\hat D(p)| < \tau \\
0 & \mbox{otherwise}
\end{cases}\]
where given a view, $D$ and $\hat D$ denote the depth map of the ground-truth and reconstructed meshes, $V$ and $\hat V$ denote the pixel visibility masks, and $\tau$ is the misalignment tolerance. A higher VSA indicates a better match between depth maps.

We evaluated the average VSA metric across 200 testing views of the NeRF-Synthetic dataset, with a tolerance set at 0.05, denoted as VSA(0.05). The results are presented in Tab.~\ref{tab:sota}. The outcomes for different misalignment tolerances are shown in Appendix Fig. 11. The comparisons clearly demonstrate that our method consistently outperforms all mesh-based methods in VSA performance, even when their meshes exhibit inferior properties compared to ours. Additionally, we provide a visual comparison of the reconstructed meshes in Fig.~\ref{fig:mesh_comparison}. Here, it is evident that our method more accurately captures intricate structures, such as the details of the ship and the legs of the drum, surpassing even nerf2mesh~\cite{tang2022nerf2mesh}, which generates non-manifold meshes (we have marked triangles with incorrect orientations in black).



\subsection{Comparison on Novel-View Synthesis}\label{sec:sota}
We present a quantitative comparison of novel view synthesis performance between our method and other neural rendering and differentiable rasterization methods on the NeRF-Synthetic dataset in Tab.~\ref{tab:sota}. We observe that while TensoRF and iNeuS achieve very high quality with their original volume rendering, their performance significantly declines when adapted to surface rendering without fine-tuning. Nvdiffrec is capable of generating watertight and manifold meshes without its final fine-tuning stage, but its rendering quality falls notably short compared to other neural rendering methods. In contrast, our models not only maintain high rendering quality but also preserve favorable mesh properties. It is noteworthy that our method achieves the highest rendering quality among all the surface rendering techniques compared, even outperforming those that generate non-manifold meshes. Moreover, by fine-tuning the mesh vertices, we can further enhance our visual fidelity, achieving an average PSNR that is 0.65 dB higher than that of the vanilla NeRF.

\input{tables/unbounded.tex}

We also provide visual comparisons of mesh-based rendering in Fig.~\ref{fig:sota}. It is evident that mesh renderings using meshes directly extracted from TensoRF and iNeuS struggle to capture high-frequency details and thin structures in the scene. This issue arises because these methods employ volume rendering during optimization, integrating the colors of multiple points along a ray to match training images. Consequently, simply extracting the color of a single point at the isosurface does not accurately reproduce the scene's appearance. While Nvdiffrec uses direct mesh rendering during training, the recovered meshes often miss complex structures, leading to a decline in visual quality. In contrast, our method, leveraging the initial training from neural volume rendering, is more adept at capturing fine-grained details of the scene. This advantage is reflected in the improved visual quality of our rendered images.

We further validate the efficacy of our method on two real-world datasets: the MipNeRF-360 dataset~\cite{barron2022mip} and the LLFF dataset~\cite{mildenhall2019local}. The quantitative results on the MipNeRF-360 dataset are presented in Tab.~\ref{tab:unbounded}, where our method notably surpasses other mesh-based methods in indoor scene reconstructions. Visual results are available in Appendix Fig. 5 and Fig. 9.

\begin{figure*}[ht]
\begin{center}
   \includegraphics[width=0.95\linewidth]{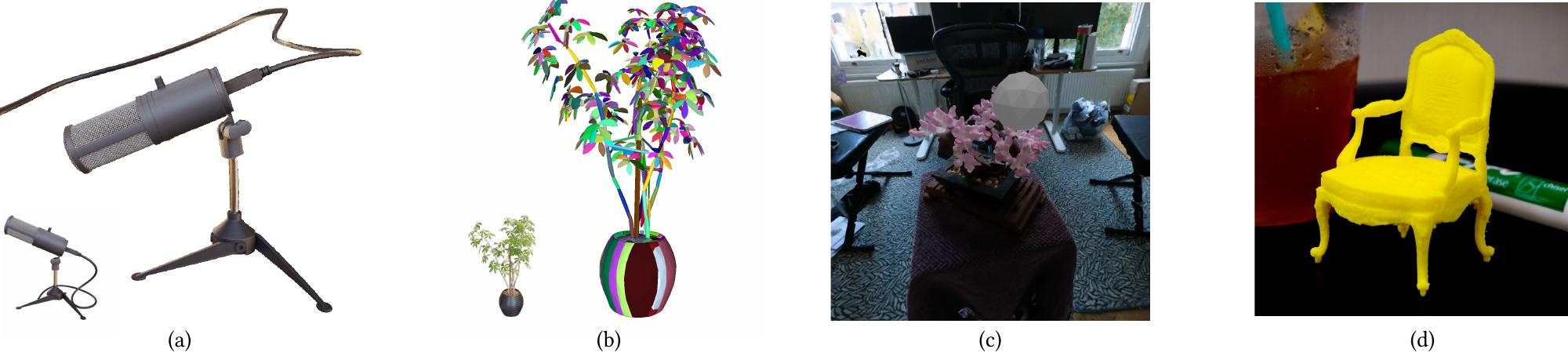}
\end{center}
\vspace{-2em}
  \caption{Applications of \neumanifold. (a) Laplacian surface editing. (b) Collision-aware convex decomposition. (c) Soft-body simulations. (d) 3D printing. }
\label{fig:application}
\vspace{-1.5em}
\end{figure*}

\input{tables/stage1_ablate.tex}

\subsection{Ablation for Stage 1 \& 2}\label{sec:stage1}
In this section, we conduct ablation studies on the NeRF-Synthetic dataset for each stage in our pipeline and also show the speed and memory cost of our DiffMC.

We begin by demonstrating that initialization from volume rendering is crucial for effective mesh optimization, with different initialization approaches yielding varying outcomes. As evidenced by the data in Tab.~\ref{tab:stage1}, optimizing the mesh without any prior initialization from volume rendering results in the poorest performance in terms of both rendering and mesh quality. Both geometry and appearance initialization enhance model performance, but geometry initialization is more pivotal for improving performance. The visual results in the Appendix also underscore this point.

We further validate the necessity of stage two in our pipeline, which focuses on the crucial role of topology optimization via differentiable iso-surface extraction; details on appearance networks are in the Appendix Tab. 7. Besides the naive way of transferring volumetric methods to surface rendering (TensoRF+MC, iNeuS+MC), another strategy involves using non-differentiable algorithms to extract meshes from the density field, then freezing the geometry while only optimizing the appearance network. While this approach can maintain good mesh properties, it results in unmodifiable topology. Our experiments reveal a significant drop in both rendering and mesh quality without topology optimization, with a notable gap of up to 2.74 PSNR as shown in Tab.~\ref{tab:stage2}.

\input{tables/diffmc_speed}
We compare the speed and GPU memory usage of our DiffMC implementation with two previous methods, DMTet~\cite{shen2021deep} and FlexiCubes~\cite{shen2023flexible} across two scenarios: simple rounded cube SDF and randomly initialized SDF. The algorithms were rigorously tested on an NVIDIA RTX 4090. Each algorithm underwent 100 repeated runs. Tab.~\ref{tab:diffmc_speed} presents the time and CUDA memory consumption for \emph{a single run}, demonstrating that our method is more efficient in terms of both memory and speed.

\subsection{Deployment Speed}\label{sec:real_time}
We present the model performance and speed after deployment in GLSL in 
Appendix Tab. 6
illustrating the trade-off between model capacity and inference speed. The two versions of our method differ only in their appearance networks: the fast version substitutes the original TensoRF MLP with Spherical Harmonics (SH), where learned features (considered as SH coefficients) are fed into a fixed SH function to decode colors, eliminating the need for any neural decoders. Both models achieve real-time rendering, with FPS based on the average time of the first frame from the NeRF-Synthetic test set on an NVIDIA RTX 4090.

%% file: tables/unbounded.tex
\begin{table}
\centering
\small
\caption{\textbf{Average PSNR on Mip-NeRF360 Scenes.} * The results for BakedSDF are sourced from its original paper and its mesh resolution is significantly higher than ours, as no code has been released. Additional metrics can be found in the Appendix.}
\label{tab:unbounded}
\vspace{-1em}
\begin{tabular}{l|c|cc}
\toprule
 Method & Geometry & Outdoor & Indoor \\
\midrule
NeRF~\cite{mildenhall2021nerf} & Volume & 21.46 & 26.84 \\
NeRF++~\cite{zhang2020nerf++} & Volume & 22.76 & 28.05 \\
mip-NeRF~\cite{barron2021mip} & Volume & 24.47 & 31.72 \\
\midrule
Mobile-NeRF~\cite{chen2022mobilenerf} & Mesh & 21.95 & - \\
BakedSDF*~\cite{yariv2023bakedsdf} & Mesh & \textbf{22.47} & \underline{27.06} \\
\midrule
Ours & Mesh & 21.07 & 25.80 \\
Ours (w/ ft)  & Mesh & \underline{22.05} & \textbf{27.63} \\
\bottomrule
\end{tabular}
\vspace{-0.5em}
\end{table}

%% file: tables/stage1_ablate.tex
\setlength{\tabcolsep}{5pt}
\begin{table}
\centering
\small
\caption{\textbf{Ablation study for Stage 1.} Using initializations from volume rendering enables more
accurate mesh reconstruction and rendering, leading to more accurate novel view synthesis.}
\label{tab:stage1}
\vspace{-1em}
\begin{tabular}[t]{c|c|ccc|c}
\toprule
      G. Init & A. Init & PSNR$\uparrow$ & SSIM$\uparrow$ & LPIPS$\downarrow$ & VSA(0.05)$\uparrow$ \\
      \midrule
      \xmark & \xmark & 20.56 & 0.826 & 0.204 & 0.214 \\
      \xmark & \cmark & 24.43 & 0.882 & 0.149 & 0.385 \\
      \cmark & \xmark & 29.74 & 0.945 & 0.067 & \textbf{0.893} \\
      \cmark & \cmark & \textbf{31.19} & \textbf{0.954} & \textbf{0.059} & 0.890\\
      \bottomrule
    \end{tabular}
\vspace{-0.5em}
\end{table} 

\setlength{\tabcolsep}{4pt}
\begin{table}
\centering
\small
\caption{\textbf{Ablation study for topology optimization in Stage 2.} Ablation on the appearance optimization is in the Appendix. }
    \label{tab:stage2}
\vspace{-1em}
\begin{tabular}[t]{l|ccc|c}
\toprule
       & PSNR$\uparrow$ & SSIM$\uparrow$ & LPIPS$\downarrow$ & VSA(0.05)$\uparrow$\\
      \midrule
      w/o Topology Opt. & 27.00 & 0.929 & 0.081 & 0.827 \\
      w/ Topology Opt. & \textbf{29.74} & \textbf{0.945} & \textbf{0.067} & \textbf{0.890} \\
      \bottomrule
    \end{tabular}
\vspace{-1em}
\end{table}

%% file: tables/diffmc_speed.tex
\setlength{\tabcolsep}{0.8pt}
\begin{table}
\small
\centering
\caption{Speed and memory comparison between DiffMC and other differentiable iso-surface extraction methods.}
\vspace{-1em}
\label{tab:diffmc_speed}
\begin{tabular}{l|cccc|cccc} 
\toprule
           & \multicolumn{4}{c|}{Rounded Cube}       & \multicolumn{4}{c}{Random Init.}                   \\ 
\midrule
           & \# V     & \# F     & Mem/G & T/ms & \# V       & \# F       & Mem/G & T/ms  \\ 
\midrule
DMTet      & 19k & 39k & 1.57       & 9.61      & 2.5M & 4.7M & 3.07       & 49.10       \\ 
FlexiCubes & 19k & 38k & 5.4        & 10.00        & 2.7M & 4.3M & 4.07       & 65.35      \\ 
DiffMC (Ours)    & 19k & 39k & 0.6        & 1.54      & 2.6M & 4.7M & 0.59       & 2.55       \\
\bottomrule
\end{tabular}
\vspace{-1.5em}
\end{table}

%% file: sections/app.tex
\methodname{} generates high-quality meshes, enabling seamless integration into graphics pipelines—a significant improvement over prior neural reconstruction methods.

\boldstart{Geometry Editing. }Effective geometry editing algorithms typically depend on reliable input mesh connectivity. As illustrated in Fig.~\ref{fig:application}a, we apply Laplacian surface editing~\cite{sorkine2004laplacian} for the non-rigid deformation of a microphone reconstructed using \methodname{}.

\boldstart{Physical Simulation. }
Our reconstructed meshes can be used as direct input to the collision-aware convex decomposition algorithm~\cite{wei2022coacd} for rigid-body collision shape generation (Fig.~\ref{fig:application}b). They can be directly converted to finite-element meshes by Delaunay tetrahedralizations~\cite{hang2015tetgen} and used in a finite-element simulation with incremental potential contact (IPC)~\cite{Li2020IPC} (Fig.~\ref{fig:teaser} and Fig.~\ref{fig:application}c).

\boldstart{3D printing. }
3D printing imposes stringent requirements on mesh quality, as it relies on slicing software to convert 3D models into printable instructions. Non-manifold models can complicate the slicing process, potentially leading to interpretation challenges for the printer. The meshes generated by \methodname{} meet these rigorous standards, enabling them to be printed as realistic objects, as demonstrated in Fig.~\ref{fig:application}d.

%% file: sections/conclusion.tex
We have introduced a novel method capable of reconstructing high-quality, watertight manifold meshes and enablinng real-time photorealistic rendering. However, our method encounters limitations in handling specular areas, as observed with the ``materials'' in NeRF-Synthetic and the ``room'' in the LLFF dataset. In such instances, the reconstructed meshes may exhibit discontinuities, reflecting the challenge of capturing different color perceptions of the same point from various viewpoints. To address this issue, we believe it will be necessary to integrate inverse rendering techniques and incorporate additional priors, aiming to achieve a more precise geometry representation

%% file: sections/supp.tex
\section{Prelimiaries}
\input{sections/prelim}

\section{Implementation Details}
For the first stage, we directly build on off-the-shelf volume rendering models. 
Specifically, for TensoRF, we use the official implementation. We compare two of our models for our main results: a high-quality one, labeled with Ours, which uses the TensoRF (VM) with 48-dim input features and 12-dim output features, plus a three-layer MLP decoder; a fast one, labeled with Ours-F that uses the TensoRF (VM) with 48-dim input features and output 27-dim SH coefficients. 

To adapt the density values for DiffMC, we transform these values into opacity using the formula: $\alpha = 1-\exp(-\sigma \cdot \delta)$, where $\sigma$ represents density, $\alpha$ denotes opacity, and $\delta$ is the ray step size used in volume rendering. We employ a threshold $t$ to control the surface's position relative to opacity and use the value $\alpha-t$ for mesh extraction.

We train all the stage 2 and 3 models with batch size of 2 for 10k iterations. We use DiffMC with a grid resolution of 256 for all results. Except when comparing with nvdiffrec, we use the default resolution of 128 as nvdiffrec's performance drops on higher resolutions, possibly due to the decreased batch size and harder optimization.

\section{Differentiable Marching Cubes (DiffMC)}\label{sec:supp_diffmc}

\begin{figure}[htbp]
    \centering
    \includegraphics[width=\linewidth]{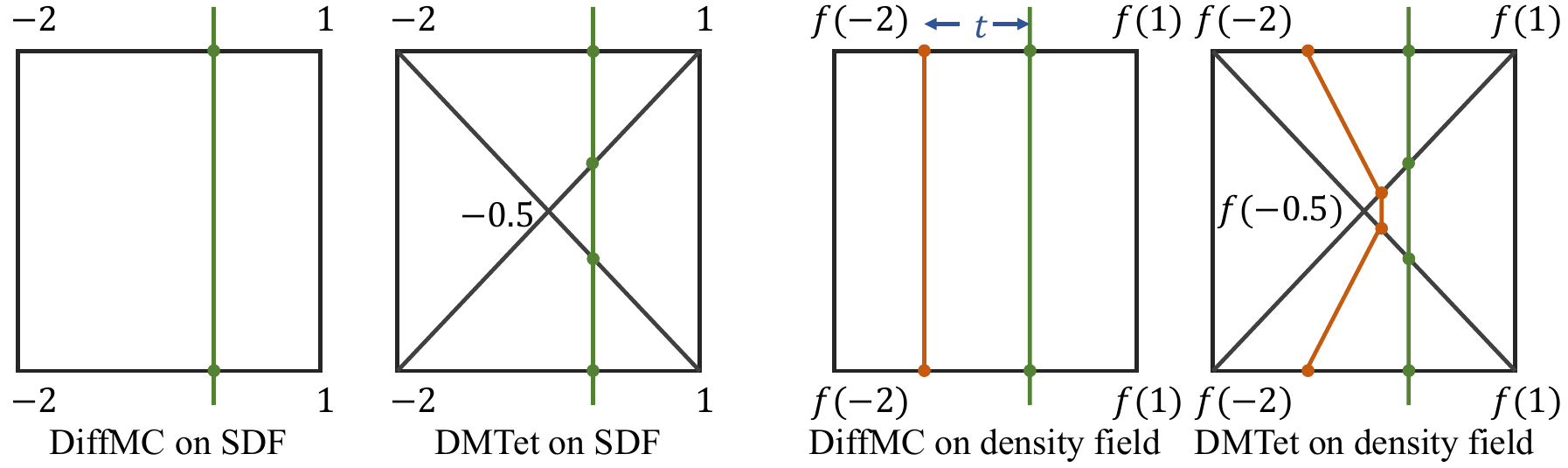}
    \vspace{-1.5em}
    \caption{2D example illustrating why DMTet tends to introduce more artifacts when extracting meshes from density fields while DiffMC can generate much smoother surfaces.}
\label{fig:supp_2d_diffmc}
\end{figure}

In this section, we present additional results for DiffMC. These include a 2D example showing why DMTet tends to introduce more artifacts on density fields than DiffMC, an ablation study that demonstrates how grid resolution influences visual fidelity and a comparison highlighting the effectiveness of our method in mesh reconstruction when compared to DMTet~\cite{shen2021deep}.

First, we illustrate how DMTet and DiffMC generate surfaces with a 2D schematic diagram in Fig.~\ref{fig:supp_2d_diffmc}. In 2D, Marching Cubes is analogous to ``Marching Squares'' and Marching Tetrahedra is analogous to ``Marching Triangles''. Given a surface (shown as a green vertical line) passing through the square/triangle grids (shown as black lines), suppose we have recorded the perfect signed distance function (SDF) values of the surface on the grid nodes, as shown in the two leftmost figures, regardless of how the algorithm divides the space, both methods exactly recover the ground truth surface through linear interpolation.

However, in practice, perfect SDF values are not easily obtainable, especially when the input comes from a volumetric density representation. Here, we simulate an imperfect SDF by applying a non-linear transformation $f(s) = \exp(s)-1-t$ to the SDF values.
Under this scenario, DiffMC can still generate a flat surface (red line in the second figure from the right), albeit with a slight offset $t$ which can be rectified by introducing an adjustable threshold to the grid values. In contrast, DMTet produces zigzag lines (red line in the rightmost figure) due to varying space divisions and cannot be easily fixed.



\input{tables/supp_reso.tex}

\begin{figure*}[htbp]
    \centering
    \includegraphics[width=\linewidth]{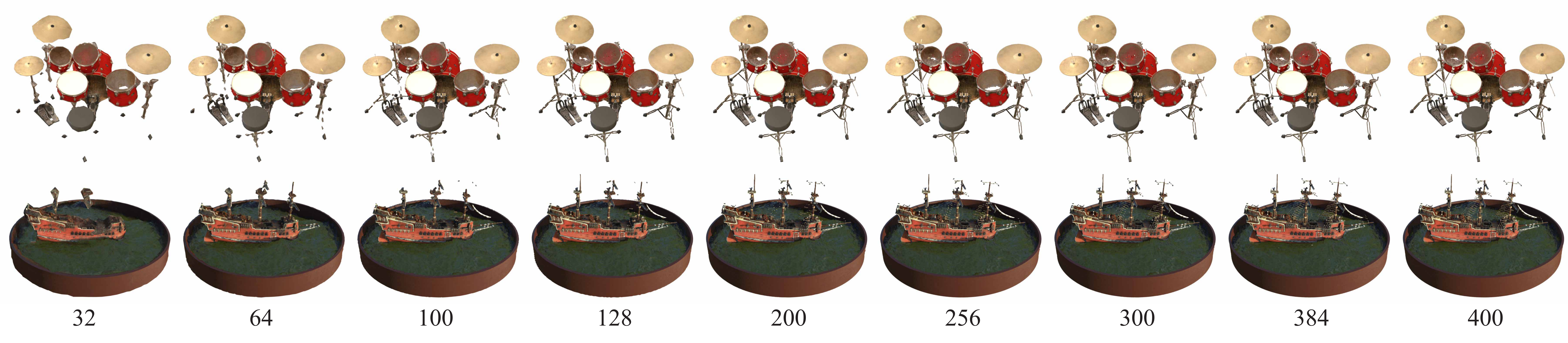}
    \vspace{-1.5em}
    \caption{The influence of DiffMC resolution to rendeirng quality. We have noticed that lower resolutions can capture most of the coarse structures but tend to lose finer details, such as the drum legs and the ropes on the ship. These finer details become more discernible as the resolution increases.}
\label{fig:supp_reso}
\end{figure*}

As we transition from lower to higher resolutions, we observe a consistent improvement in rendering quality, ultimately converging as the resolution reaches 400, as demonstrated in Tab. \ref{tab:supp_reso}. Moreover, as depicted in Fig.~\ref{fig:supp_reso}, a higher-resolution DiffMC is notably more adept at recovering intricate structures, such as the ropes on the ship.

\begin{figure*}[htbp]
    \centering
    \includegraphics[width=\linewidth]{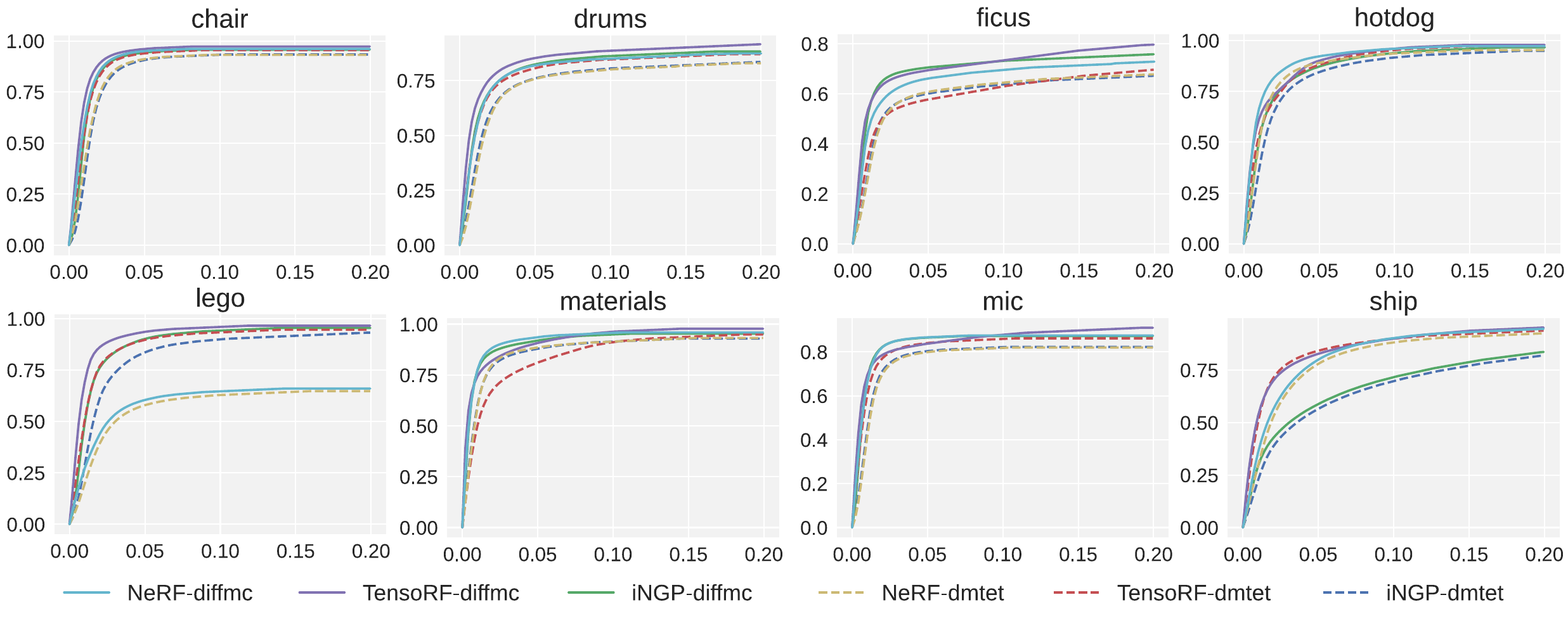}
    \vspace{-1.5em}
    \caption{DMTet vs DiffMC on extracting meshes from pre-trained density fields. Across all three methods, DiffMC consistently outperforms DMTet in terms of mesh quality.}
\label{fig:supp_dmtet}
\end{figure*}

 Next, we highlight the advantages of our method in extracting meshes from density fields by applying both our approach and DMTet~\cite{shen2021deep} to a set of pre-trained density networks, including TensoRF~\cite{chen2022tensorf}, instant-NGP~\cite{muller2022instant} and vanilla NeRF~\cite{mildenhall2021nerf}. By comparing the visible surface agreemen (VSA) of the reconstructed meshes, as illustrated in Fig.~\ref{fig:supp_dmtet}, we observe a consistent enhancement brought about by DiffMC across all methods. We also conduct a comparison between our DiffMC and DMTet in our pipeline, noting a significant improvement in surface smoothness with our method, which effectively mitigates most of the artifacts resulting from the non-linearity of the density field.

\begin{figure*}[htbp]
    \centering
    \includegraphics[width=\linewidth]{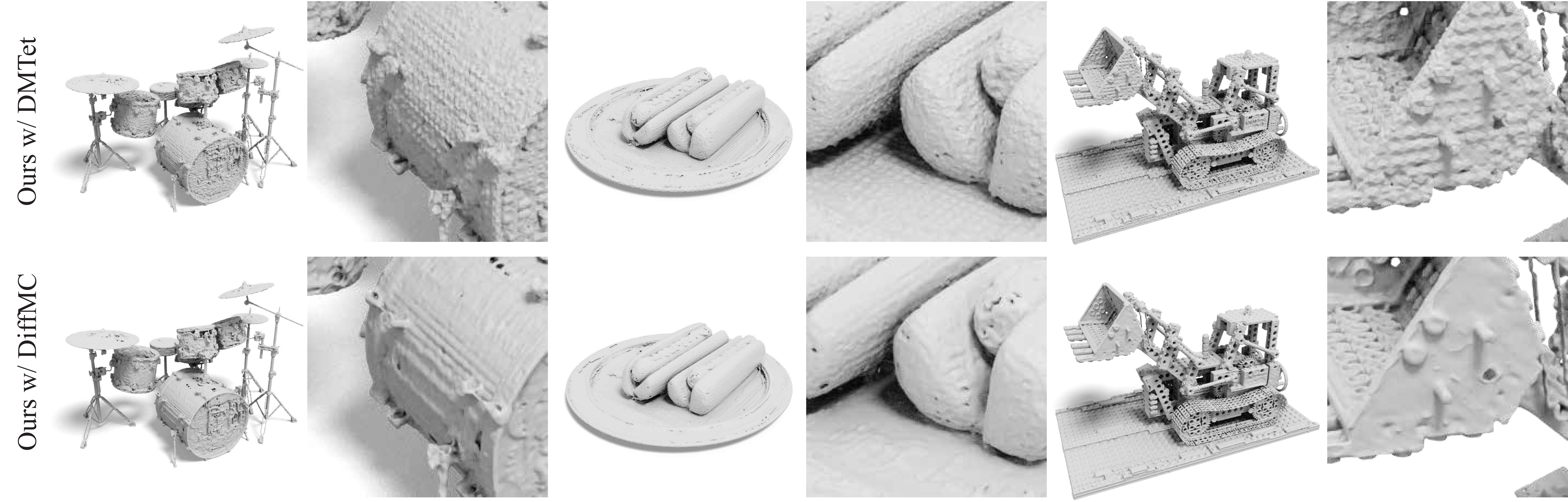}
    \vspace{-1.5em}
    \caption{A mesh surface comparison of Ours between using DiffMC and DMTet reveals that DiffMC can create significantly smoother surfaces. This improvement is not limited to axis-aligned surfaces; it consistently outperforms DMTet on various rounded surfaces as well.}
\label{fig:supp_diffmc_mesh}
\end{figure*}

\section{Mip-NeRF 360 Dataset}

We evaluate our method on unbounded real scenes in the Mip-NeRF 360 dataset~\cite{barron2022mip}. To deal with the unbounded background, we follow the contraction function proposed in ~\cite{barron2022mip} to warp the far objects from their original space, $t$-space, into the contracted space $s$-space (a sphere with a radius of 1.2 in our setup). When generating the mesh, we apply DiffMC on the geometry network within $t$-space so that the mesh can be watertight manifold, otherwise the contraction may break the property. After getting the points on the mesh surface, we contract the points back to $s$-space to compute the color. Within the $t$-space, we utilize multiple resolutions for the entire scene, with a higher resolution (340) for the foreground and a lower resolution (56) for the background. To represent the distant background that falls outside the [-4, 4] box range, we employ a skybox. We use the anti-aliasing of nvdiffrast~\cite{Laine2020diffrast} for this dataset.

\begin{figure*}[ht]
    \centering
    \includegraphics[width=\linewidth]{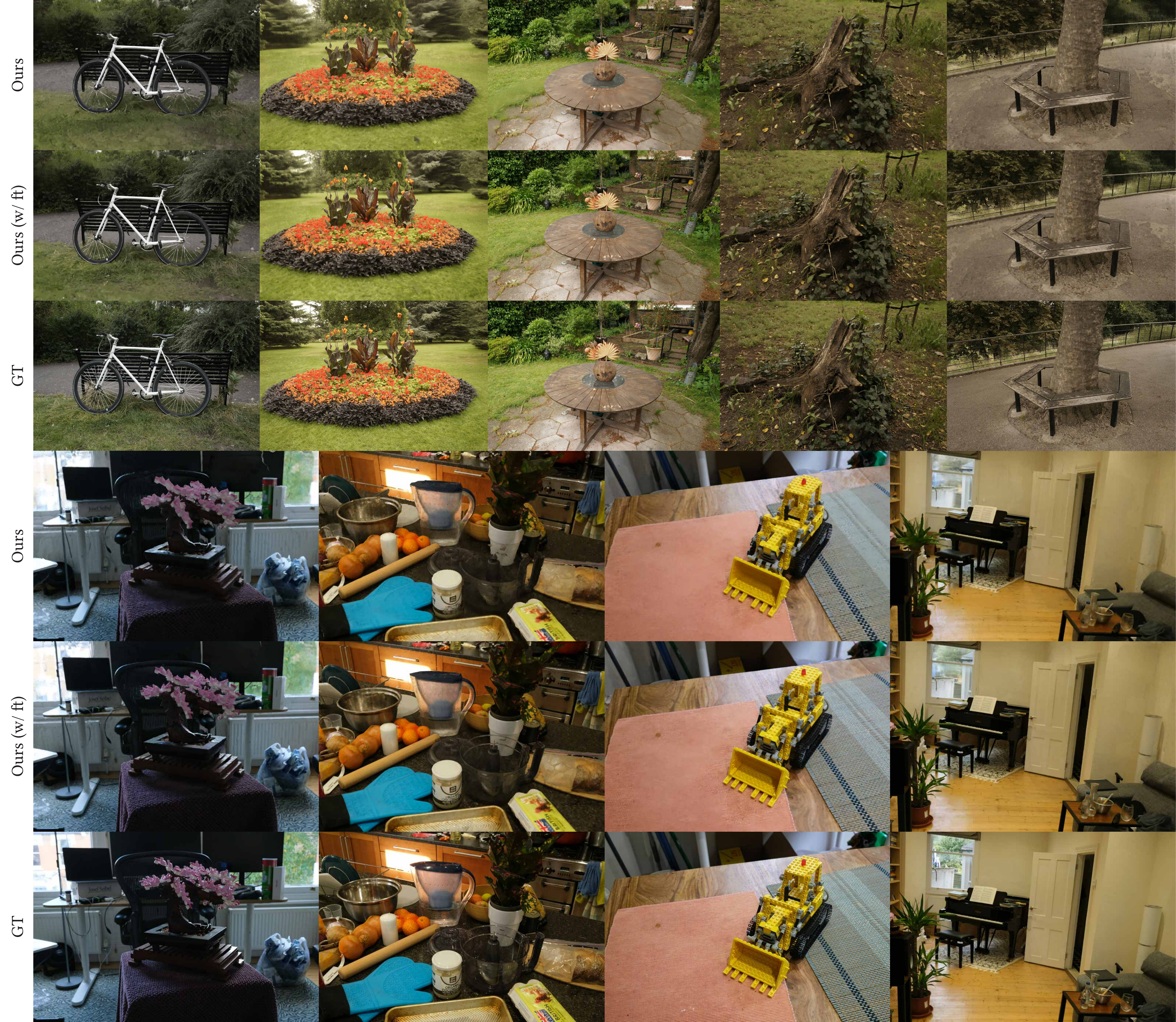}
    \vspace{-1.5em}
    \caption{Mip-NeRF 360 renderings.}
\label{fig:fig_unbounded}
\end{figure*}

\input{tables/supp_unbounded.tex}

\begin{figure}[htbp]
    \centering
    \includegraphics[width=\linewidth]{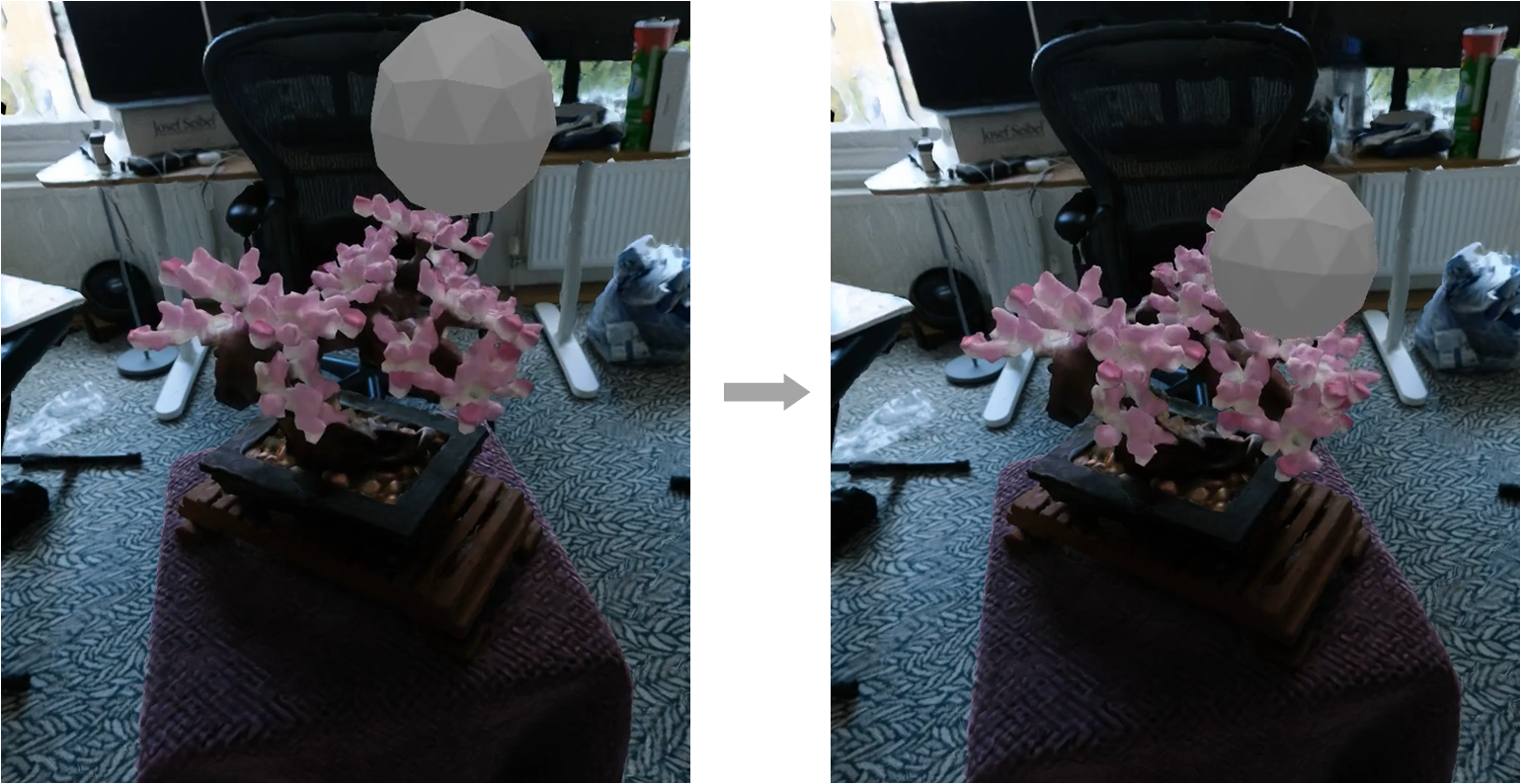}
    \caption{Soft-body simulation on the foreground watertight manifold mesh. The solid ball hits the flower and makes it deform. See the attached video.}
\label{fig:supp_flowers}
\end{figure}

Our method generates watertight manifold foreground meshes. Therefore, we can apply simulation algorithms on the foreground objects, as shown in Fig.~\ref{fig:supp_flowers}, where we apply soft-body simulation on the flower and use a solid ball to hit it.

In Tab. \ref{tab:supp_unbounded}, we compared our method with others. Some mesh rendering methods, such as MobileNeRF ~\cite{chen2022mobilenerf} and nerf2mesh ~\cite{tang2022nerf2mesh}, provided results for selected scenes, while our method worked effectively on all unbounded scenes, particularly excelling in indoor scenes. 


\begin{figure}[htbp]
    \centering
    \includegraphics[width=\linewidth]{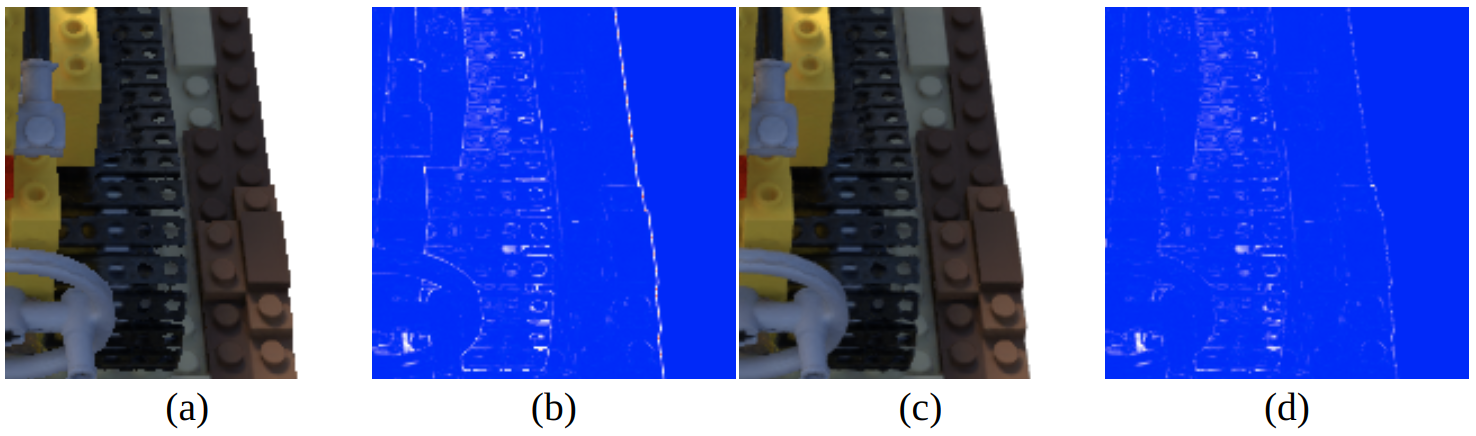}
    \caption{Comparison between 8x MSAA and no AA. (a) Our deployed high-quality model without AA (FPS: 146, PSNR: 31.26). (c) the same model with 8$\times$ MSAA (FPS: 93, PSNR: 33.01). (b) and (d) show the error maps of (a) and (c) respectively. The visual quality at edges is significantly improved by MSAA with a relatively small performance hit. }
\label{fig:supp_aa}
\end{figure}

\begin{figure*}[ht]
    \centering
    \includegraphics[width=\linewidth]{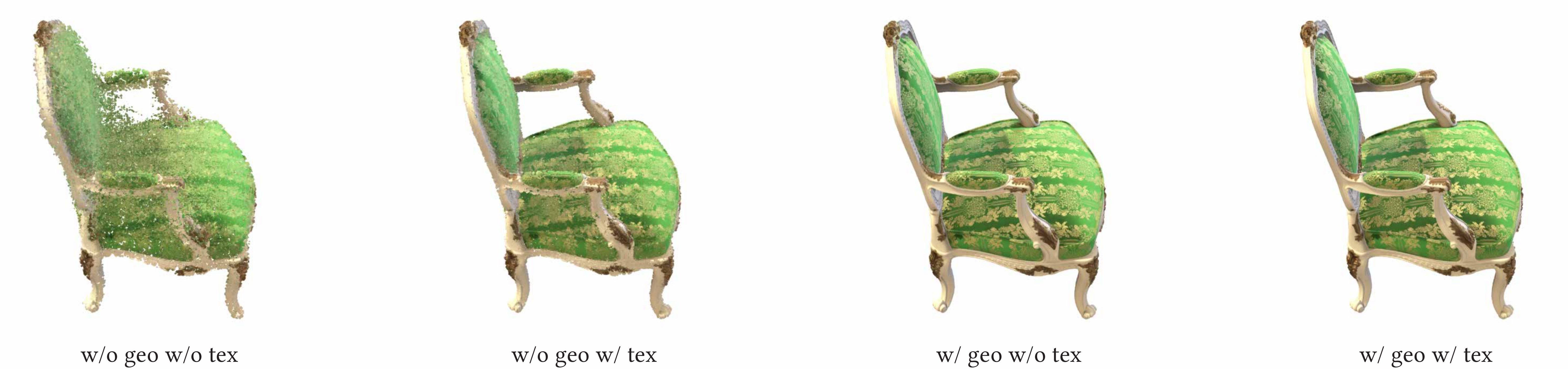}
    \caption{Visual comparison of Ours w/ or w/o geometry and texture initialization. when both initializations are omitted, the mesh optimization process can easily become trapped in local minima, as illustrated in the first left image. Although texture initialization can provide some assistance to the optimization process, it still falls short of achieving satisfactory geometric quality.}
\label{fig:supp_fig_init}
\end{figure*}

\section{LLFF dataset}

\begin{figure*}[htbp]
    \centering
    \includegraphics[width=\linewidth]{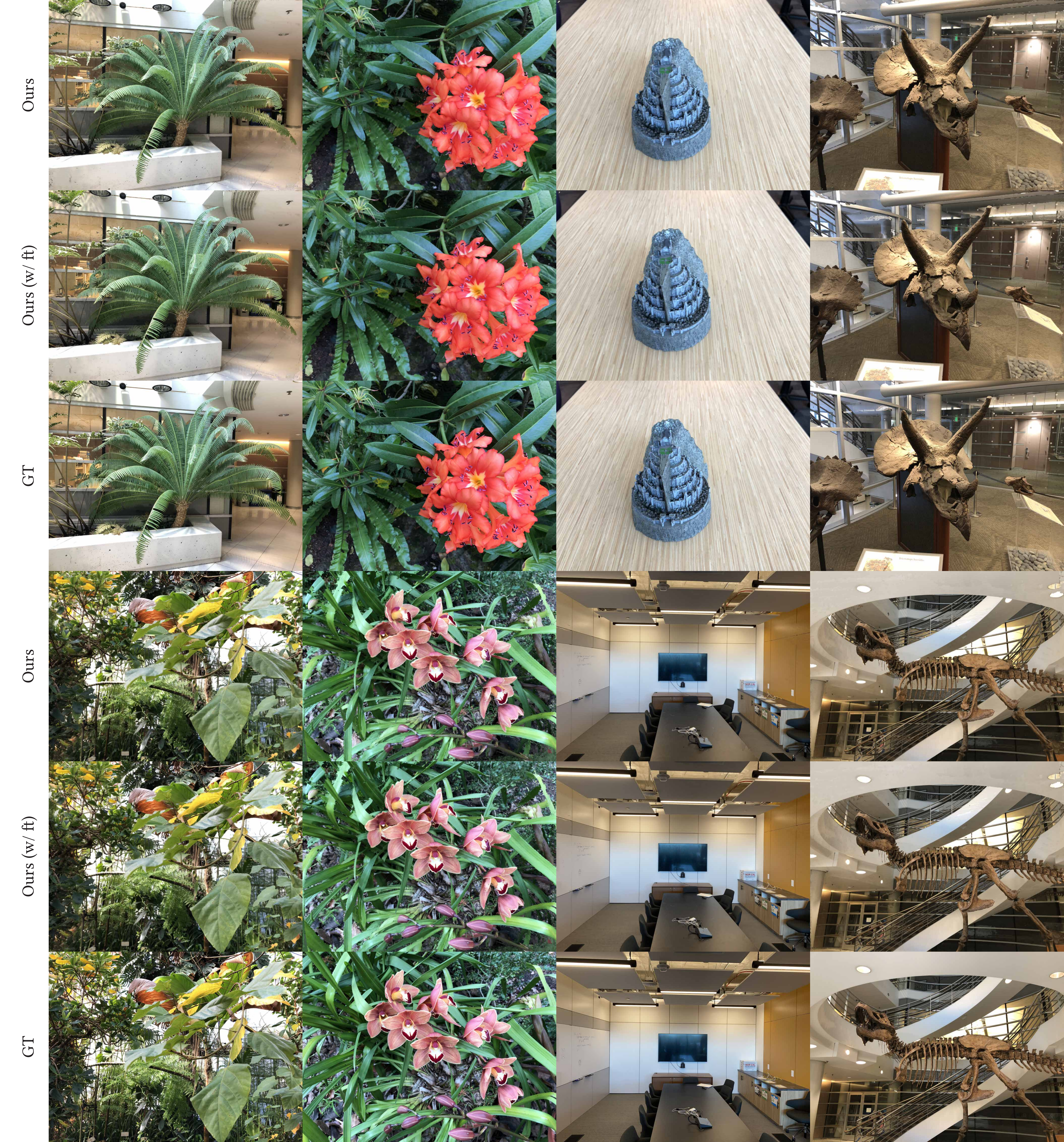}
    \vspace{-1.5em}
    \caption{LLFF renderings.}
\label{fig:supp_llff}
\end{figure*}

\begin{figure*}[htbp]
    \centering
    \includegraphics[width=\linewidth]{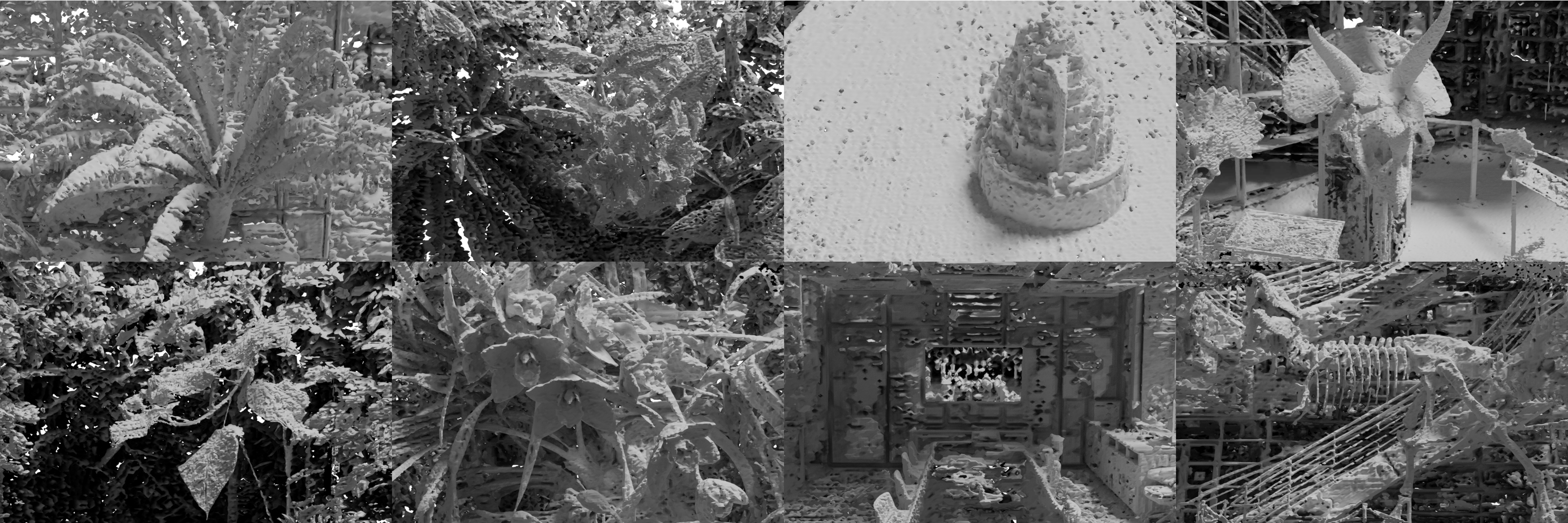}
    \vspace{-1.5em}
    \caption{LLFF mesh.}
\label{fig:supp_llff_mesh}
\end{figure*}

We evaluate our method on forward-facing scenes on LLFF dataset~\cite{mildenhall2019local}. Following ~\cite{chen2022tensorf}, we contract the whole scene into NDC space to do the reconstruction and mesh extraction. On this dataset, we use DiffMC with resolution of 375. We use 9$\times$ sample per-pixel SSAA for this dataset. Tab.~\ref{tab:supp_llff} and Fig. ~\ref{fig:supp_llff} shows the quantitative and qualitative results. Fig.~\ref{fig:supp_llff_mesh} shows the reconstructed mesh of the scenes.

We put our method to the test with forward-facing scenes from the LLFF dataset ~\cite{mildenhall2019local}. In line with ~\cite{chen2022tensorf}, we condensed the entire scene into NDC space for reconstruction and mesh extraction. For this dataset, we employed DiffMC with a resolution of 375. You can find both the quantitative results in Tab. \ref{tab:supp_llff} and the qualitative results in Fig.~\ref{fig:supp_llff}. Additionally, Fig.~\ref{fig:supp_llff_mesh} showcases the reconstructed mesh for these scenes.

\input{tables/supp_llff.tex}

\section{NeRF-Synthetic Dataset}

\input{tables/supp_syn.tex}

We show the complete quantitative comparison between our method and the previous works on the NeRF-Synthetic dataset in Tab.~\ref{tab:supp_sota_render} and the complete visual comparison in Fig.~\ref{fig:supp_syn}.

\input{tables/supp_tf.tex}


\begin{figure*}[htbp]
    \centering
    \includegraphics[width=\linewidth]{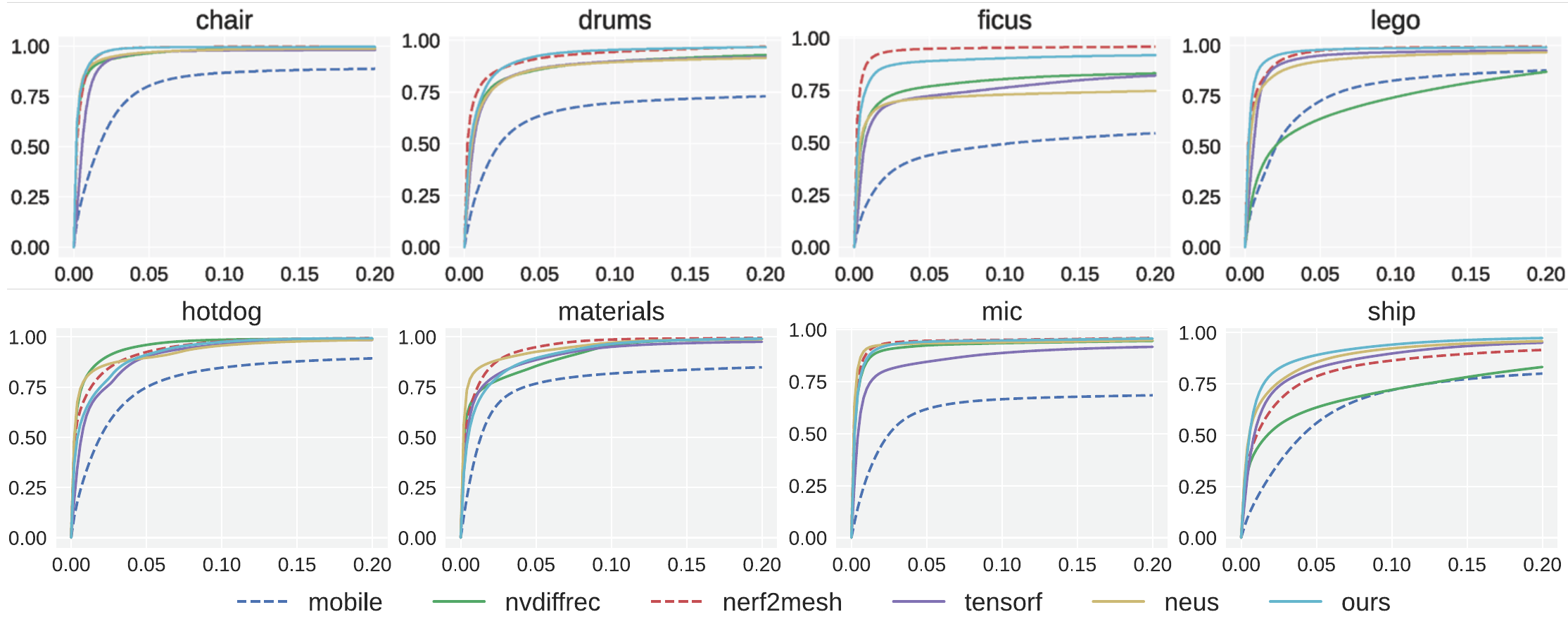}
    \caption{VSA plots for different misalignment tolerances.}
\label{fig:supp_VSA}
\end{figure*}

\begin{figure*}[htbp]
    \centering
    \includegraphics[width=0.9\linewidth]{figures/supp_syn.pdf}
    \caption{NeRF-Synthetic renderings.}
\label{fig:supp_syn}
\end{figure*}

\section{Mesh Quality}
We show the mesh quality comparison in Fig.\ref{fig:supp_syn_mesh}, where except for Mobile-NeRF~\cite{chen2022mobilenerf} and nerf2mesh~\cite{tang2022nerf2mesh}, all the meshes are watertight manifold. We show the VSA-tolerance curves for the scenes in NeRF-Synthetic in Fig.~\ref{fig:supp_VSA}.

\begin{figure*}[htbp]
    \centering
    \includegraphics[width=\linewidth]{figures/supp_syn_mesh.pdf}
    \vspace{-1.5em}
    \caption{NeRF-Synthetic mesh.}
\label{fig:supp_syn_mesh}
\end{figure*}

\section{Network Architecture}
In this section, we describe the network architecture used in the experiments. Our proposed method has two versions, a high-quality one and a fast one, and they share the same geometry network architecture but with different appearance networks. The geometry network is the same as TensoRF~\cite{chen2022tensorf} VM-192 in its paper. The appearance network is from TensoRF and we show the two versions below respectively.

\paragraph{High-quality.}

We use the Vector-Matrix (VM) decomposition in TensoRF, which factorizes a tensor into multiple vectors and matrices along the axes as in Equation 3 of the TensoRF paper. The feature $\mathcal{G}_c(\mathbf{x})$ generated by VM decomposition is concatenated with the viewing direction $d$ and put into the MLP decoder $S$ for the output color $c$:

\begin{equation}
    c = S(\mathcal{G}_c(\mathbf{x}),d),
\end{equation}
We also apply frequency encodings (with Sin and Cos functions) on both the features $\mathcal{G}_c(\mathbf{x})$ and the viewing direction $d$. We use a $300^3$ dense grid to represent the scenes in NeRF-Synthetic and use 2 frequencies for features and 6 frequencies for the viewing direction. The detailed network architecture is shown in Tab.~\ref{tab:supp_tensorf}. As for Mip-NeRF 360 and LLFF datasets we use a $512^3$ dense grid to represent the unbounded indoor scenes and do not use frequency encodings.

\paragraph{Fast.}
The fast version shares similar architecture and positional encoding setups with the high-quality version before the MLP decoder but uses the spherical harmonics (SH) function as $\mathcal{G}_c$ instead, as shown in Tab.~\ref{tab:supp_tensorf}.

\paragraph{Quality Speed Trade-off}
\input{tables/supp_quality_speed}
We also show the model rendering quality and speed after deployment in Tab.~\ref{tab:trade-off}.



\section{Visualization of ablation Study on Stage 1}
We visualize the rendering results of models using different initialization strategies during stage 1, as shown in Fig.~\ref{fig:supp_fig_init}. The comparison shows that employing high-resolution DiffMC grids without proper geometry initialization can lead the mesh optimization process to become stuck in suboptimal geometric configurations.

\section{Ablation Study on Appearance Network}
\input{tables/supp_app}
We validate the necessity of optimizing the meshes in Tab.~\ref{tab:supp_app}. To achieve this, we compare against
baselines that keep the meshes from Stage 1 fixed and only optimize the appearance. We also provide
the results using the GT mesh in combination with the TensorF appearance network as a reference,
representing the upper limit of texture optimization methods. As we can see from the results, TensoRF appearance network achieves the best performance. All
appearance networks were trained from scratch for fair comparison.

%% file: sections/prelim.tex
\subsection{Watertight and Manifold Meshes}
\boldstart{Watertight. } If all edges are shared by exactly two faces, then the mesh is watertight.

\boldstartspace{Manifold. } A manifold mesh must meet the following properties: (1) all edges must connect at most two faces; (2) each edge is incident to one or two faces and faces incident to a vertex must form a closed or open fan; (3) the faces must not self-intersect with each other.

\subsection{Volumetric Neural Fields}
Recent neural field representations utilize differentiable volume rendering for their reconstruction and leads to high visual quality.
While our approach can generally support any neural field models, we apply TensoRF and NeuS in our paper. We now briefly cover the preliminaries of the method.

The original NeRF uses pure MLPs, which make it slow to train and incapable of modeling details accurately.
TensoRF~\cite{chen2022tensorf} decodes the radiance field from a volume of features, and this feature volume is further factorized into factors leveraging CANDECOMP/PARAFAC decomposition or vector-matrix decomposition. In this work, we are interested in the vector-matrix decomposition, which factorizes the 4D feature volume as the sum of three outer products between a matrix and a vector. 




\subsection{Differentiable Rasterization}
\label{sec:diffrast}
Differentiable rasterization refers to methods that optimize inputs of rasterization-based rendering pipelines. In this work, we are interested in nvdiffrast~\cite{Laine2020diffrast}, which consists of 4 stages, rasterization, interpolation, texture lookup, and anti-aliasing. We mainly use the rasterization stage, which maps triangles from 3D space onto pixel space, and the interpolation stage, which provides 3D coordinates of pixels to query the appearance network.

To ensure the mesh optimized by differentiable rasterization is a watertight manifold, we need to apply a meshing algorithm that generates such meshes. In this work we propose DiffMC, which divides the 3D space into a deformable grid and takes a scalar field (often SDF) defined on its vertices as input. The algorithm turns the scalar field into an explicit mesh by a differentiable marching cubes algorithm.

%% file: tables/supp_reso.tex
\begin{table*}[!ht]
    \centering
    \caption{The influence of DiffMC resolution to rendering quality. The visual fidelity consistently improves as the resolution increases, eventually reaching a plateau when it reaches 400.}
\label{tab:supp_reso}
    \begin{tabular}{l|ccccccccc}
    \toprule
        DiffMC reso & 32 & 64 & 100 & 128 & 200 & 256 & 300 & 384 & 400 \\ \hline
        PSNR & 23.12 & 26.83 & 28.64 & 29.46 & 30.8 & 31.19 & 31.34 & 31.53 & 31.54 \\ \hline
        SSIM & 0.894 & 0.925 & 0.94 & 0.946 & 0.952 & 0.954 & 0.955 & 0.956 & 0.956 \\ \hline
        LPIPS & 0.121 & 0.089 & 0.075 & 0.069 & 0.061 & 0.059 & 0.057 & 0.056 & 0.056 \\ 
    \bottomrule
    \end{tabular}
\end{table*}

%% file: tables/supp_unbounded.tex
\begin{table*}[!ht]
    \centering
    \small
    \caption{Quantitative results on each scene in the Mip-NeRF 360 dataset.}
\label{tab:supp_unbounded}
    \begin{tabular}{l|ccccc|cccc|c}
    \toprule
        PSNR & Bicycle & Garden & Stump & Flowers & Treehill & Bonsai & Counter & Kitchen & Room & Mean \\ \hline
        MobileNeRF & \textbf{21.70} & 23.54 & \textbf{23.95} & \textbf{18.86} & \textbf{21.72} & - & - & - & - & - \\ 
        nerf2mesh & 22.16 & 22.39 & 22.53 & - & - & - & - & - & - & - \\ 
        BakedSDF & - & - & - & - & - & - & - & - & - & 24.51 \\ \hline
        Ours (HQ-m) & 20.16 & 23.36 & 22.27 & 18.49 & 21.07 & 26.64 & 24.83 & 24.97 & 26.75 & 23.17 \\ 
        Ours (HQ) & 21.38 & \textbf{24.90} & 23.51 & 18.82 & 21.64 & \textbf{28.61} & \textbf{26.31} & \textbf{26.63} & \textbf{28.95} & \textbf{24.53} \\ \hline \hline
        SSIM & ~ & ~ & ~ & ~ & ~ & ~ & ~ & ~ & ~ & ~ \\ \hline
        MobileNeRF & 0.426 & 0.599 & 0.556 & 0.321 & 0.450 & - & - & - & - & - \\ 
        nerf2mesh & \textbf{0.470} & 0.500 & 0.508 & - & - & - & - & - & - & - \\
        BakedSDF & - & - & - & - & - & - & - & - & - & \textbf{0.697} \\ \hline
        Ours (HQ-m) & 0.382 & 0.616 & 0.492 & 0.334 & 0.447 & 0.835 & 0.746 & 0.644 & 0.815 & 0.590 \\
        Ours (HQ) & 0.469 & \textbf{0.746} & \textbf{0.589} & \textbf{0.366} & \textbf{0.494} & \textbf{0.888} & \textbf{0.808} & \textbf{0.764} & \textbf{0.872} & 0.666 \\ \hline \hline
        LPIPS & ~ & ~ & ~ & ~ & ~ & ~ & ~ & ~ & ~ & ~ \\ \hline
        MobileNeRF & 0.513 & 0.358 & 0.430 & 0.526 & 0.522 & - & - & - & - & - \\
        nerf2mesh & 0.510 & 0.434 & 0.490 & - & - & - & - & - & - & - \\
        BakedSDF & - & - & - & - & - & - & - & - & - & \textbf{0.309} \\ \hline
        Ours (HQ-m) & 0.561 & 0.372 & 0.475 & 0.553 & 0.560 & 0.268 & 0.346 & 0.380 & 0.348 & 0.429 \\
        Ours (HQ) & \textbf{0.488} & \textbf{0.252} & \textbf{0.413} & \textbf{0.520} & \textbf{0.506} & \textbf{0.201} & \textbf{0.270} & \textbf{0.275} & \textbf{0.274} & 0.355 \\
    \bottomrule
    \end{tabular}
\end{table*}

%% file: tables/supp_llff.tex
\begin{table*}[!ht]
    \centering
    \caption{Quantitative results on each scene in the LLFF dataset.}
\label{tab:supp_llff}
    \begin{tabular}{l|cccccccc|c}
    \toprule
        PSNR & Fern & Flower & Fortress & Horns & Leaves & Orchids & Room & Trex & Mean \\ \hline
        MobileNeRF & \textbf{24.59} & 27.05 & 30.82 & 27.09 & 20.54 & 19.66 & \textbf{31.28} & 26.26 & 25.91 \\ 
        nerf2mesh & 23.94 & 26.48 & 28.02 & 26.25 & 19.22 & 19.08 & 29.24 & 25.80 & 24.75 \\ \hline
        Ours (F-m) & 23.72 & 27.05 & 30.88 & 27.01 & 19.68 & 18.43 & 30.33 & 25.03 & 25.27 \\ 
        Ours (F) & 24.05 & \textbf{27.22} & 30.98 & 27.09 & 19.92 & 18.91 & 30.63 & 25.58 & 25.55 \\ 
        Ours (HQ-m) & 24.19 & 26.99 & 31.18 & 27.35 & 20.49 & 19.68 & 30.79 & 26.61 & 25.91 \\ 
        Ours (HQ) & 24.54 & 27.08 & \textbf{31.32} & \textbf{27.49} & \textbf{20.59} & \textbf{19.73} & 31.11 & \textbf{27.16} & \textbf{26.13} \\ \hline \hline
        SSIM & ~ & ~ & ~ & ~ & ~ & ~ & ~ & ~ & ~ \\ \hline
        MobileNeRF & \textbf{0.808} & 0.839 & 0.891 & 0.864 & 0.711 & 0.647 & \textbf{0.943} & 0.900 & 0.825 \\ 
        nerf2mesh & 0.751 & \textbf{0.879} & 0.765 & 0.819 & 0.644 & 0.602 & 0.914 & 0.868 & 0.780 \\ \hline
        Ours (F-m) & 0.757 & 0.842 & 0.895 & 0.864 & 0.681 & 0.601 & 0.923 & 0.865 & 0.803 \\ 
        Ours (F) & 0.772 & 0.848 & 0.898 & 0.866 & 0.693 & 0.622 & 0.926 & 0.872 & 0.812 \\ 
        Ours (HQ-m) & 0.789 & 0.852 & \textbf{0.902} & 0.877 & 0.739 & 0.677 & 0.930 & 0.896 & 0.833 \\ 
        Ours (HQ) & 0.801 & 0.856 & \textbf{0.902} & \textbf{0.881} & \textbf{0.745} & \textbf{0.681} & 0.933 & \textbf{0.904} & \textbf{0.838} \\ \hline \hline
        LPIPS & ~ & ~ & ~ & ~ & ~ & ~ & ~ & ~ & ~ \\ \hline
        MobileNeRF & \textbf{0.202} & 0.163 & \textbf{0.115} & 0.169 & 0.245 & 0.277 & \textbf{0.143} & \textbf{0.147} & \textbf{0.183} \\ 
        nerf2mesh & 0.303 & 0.204 & 0.270 & 0.260 & 0.321 & 0.314 & 0.246 & 0.215 & 0.267 \\ \hline
        Ours (F-m) & 0.274 & 0.181 & 0.158 & 0.196 & 0.254 & 0.278 & 0.208 & 0.256 & 0.226 \\ 
        Ours (F) & 0.258 & 0.175 & 0.152 & 0.191 & 0.244 & 0.260 & 0.203 & 0.247 & 0.216 \\ 
        Ours (HQ-m) & 0.245 & 0.164 & 0.137 & 0.171 & 0.202 & 0.234 & 0.188 & 0.216 & 0.195 \\ 
        Ours (HQ) & 0.228 & \textbf{0.160} & 0.136 & \textbf{0.165} & \textbf{0.198} & \textbf{0.226} & 0.181 & 0.205 & 0.187 \\ 
    \bottomrule
    \end{tabular}
\end{table*}

%% file: tables/supp_syn.tex
\begin{table*}[!ht]
    \centering
    \caption{Quantitative results on each scene in the NeRF-Synthetic dataset.}
\label{tab:supp_sota_render}
    \begin{tabular}{l|cccccccc|c}
    \toprule
        PSNR & Chair & Drums & Ficus & Hotdog & Lego & Materials & Mic & Ship & Mean \\ \hline
        MobileNeRF & 34.09 & 25.02 & 30.20 & 35.46 & 34.18 & 26.72 & 32.48 & 29.06 & 30.90 \\ 
        nvdiffrec & 31.00 & 24.39 & 29.86 & 33.27 & 29.61 & 26.64 & 30.37 & 26.05 & 28.90 \\ 
        TensoRF (DT) & 27.72 & 22.20 & 25.66 & 28.85 & 25.86 & 22.12 & 26.13 & 23.67 & 25.28 \\ 
        NeuS (DT) & 31.80 & 22.52 & 23.44 & 33.86 & 28.07 & 26.68 & 31.42 & 25.02 & 27.85 \\ 
        nerf2mesh & 31.93 & 24.80 & 29.81 & 34.11 & 32.07 & 25.45 & 31.25 & 28.69 & 29.76 \\ 
        nvdiffrec (m) & 31.24 & 23.17 & 25.11 & 32.67 & 28.44 & 26.33 & 29.39 & 24.82 & 27.65 \\ \hline
        Ours (F) & 33.82 & 25.25 & 31.28 & 35.43 & 34.40 & 26.83 & 32.37 & 28.13 & 30.94 \\ 
        Ours (HQ) & \textbf{34.46} & \textbf{25.42} & \textbf{31.83} & \textbf{36.45} & \textbf{35.40} & \textbf{27.38} & \textbf{33.46} & \textbf{28.77} & \textbf{31.65} \\ 
        Ours (F-m) & 33.68 & 24.98 & 30.23 & 35.10 & 33.39 & 26.61 & 32.21 & 27.54 & 30.47 \\ 
        Ours (HQ-m) & 34.37 & 25.17 & 30.64 & 36.35 & 34.28 & 27.22 & 33.35 & 28.12 & 31.19 \\ \hline \hline
        SSIM & ~ & ~ & ~ & ~ & ~ & ~ & ~ & ~ & ~ \\ \hline
        MobileNeRF & 0.978 & 0.927 & 0.965 & 0.973 & 0.975 & 0.913 & 0.979 & 0.867 & 0.947 \\ 
        nvdiffrec & 0.965 & 0.921 & 0.969 & 0.973 & 0.952 & 0.924 & 0.975 & 0.827 & 0.938 \\ 
        TensoRF (DT) & 0.922 & 0.872 & 0.933 & 0.916 & 0.893 & 0.835 & 0.936 & 0.780 & 0.886 \\ 
        NeuS (DT) & 0.975 & 0.907 & 0.934 & 0.975 & 0.949 & 0.921 & 0.981 & 0.840 & 0.935 \\
        nerf2mesh & 0.964 & 0.927 & 0.967 & 0.970 & 0.957 & 0.896 & 0.974 & 0.865 & 0.940 \\ 
        nvdiffrec (m) & 0.970 & 0.915 & 0.937 & 0.973 & 0.943 & 0.927 & 0.975 & 0.820 & 0.932 \\ \hline
        Ours (F) & 0.977 & 0.935 & 0.974 & 0.978 & 0.978 & 0.925 & 0.981 & 0.865 & 0.952 \\ 
        Ours (HQ) & \textbf{0.981} & \textbf{0.939} & \textbf{0.977} & \textbf{0.981} & \textbf{0.982} & \textbf{0.930} & \textbf{0.986} & \textbf{0.877} & \textbf{0.956} \\ 
        Ours (F-m) & 0.976 & 0.932 & 0.970 & 0.978 & 0.976 & 0.923 & 0.980 & 0.859 & 0.949 \\ 
        Ours (HQ-m) & \textbf{0.981} & 0.935 & 0.973 & \textbf{0.981} & 0.979 & 0.928 & 0.985 & 0.871 & 0.954 \\ \hline \hline
        LPIPS & ~ & ~ & ~ & ~ & ~ & ~ & ~ & ~ & ~ \\  \hline
        MobileNeRF & 0.025 & 0.077 & 0.048 & 0.050 & 0.025 & 0.092 & 0.032 & 0.145 & 0.062 \\ 
        nvdiffrec & 0.023 & 0.086 & 0.032 & 0.064 & 0.047 & 0.111 & 0.031 & 0.188 & 0.073 \\ 
        TensoRF (DT) & 0.076 & 0.130 & 0.070 & 0.113 & 0.090 & 0.146 & 0.070 & 0.230 & 0.115 \\ 
        NeuS (DT) & 0.033 & 0.101 & 0.065 & 0.041 & 0.056 & 0.084 & 0.021 & 0.191 & 0.074 \\ 
        nerf2mesh & 0.046 & 0.084 & 0.045 & 0.060 & 0.047 & 0.107 & 0.042 & 0.145 & 0.072 \\ 
        nvdiffrec (m) & 0.020 & 0.104 & 0.057 & 0.068 & 0.059 & 0.116 & 0.028 & 0.220 & 0.084 \\ \hline
        Ours (F) & 0.036 & 0.073 & 0.035 & 0.041 & 0.027 & 0.089 & 0.024 & 0.167 & 0.061 \\ 
        Ours (HQ) & \textbf{0.026} & \textbf{0.068} & \textbf{0.033} & \textbf{0.035} & \textbf{0.023} & \textbf{0.085} & \textbf{0.017} & \textbf{0.159} & \textbf{0.056} \\ 
        Ours (F-m) & 0.037 & 0.079 & 0.040 & 0.043 & 0.031 & 0.091 & 0.024 & 0.174 & 0.065 \\ 
        Ours (HQ-m) & 0.027 & 0.074 & 0.038 & 0.036 & 0.027 & 0.086 & \textbf{0.017} & 0.164 & 0.059 \\
    \bottomrule
    \end{tabular}
\end{table*}

%% file: tables/supp_tf.tex
\begin{table*}[ht]
    \centering
    \caption{Appearance network architecture of Ours (HQ) and Ours (F) for NeRF-Synthetic.}
\label{tab:supp_tensorf}
    \begin{tabular}{l|c|c}
    \toprule
        Name & High-Quality & Fast \\ \hline
        app matrix xy & Param (48 x 300 x 300) & Param (48 x 300 x 300) \\ \hline
        app matrix yz & Param (48 x 300 x 300) & Param (48 x 300 x 300) \\ \hline
        app matrix zx & Param (48 x 300 x 300) & Param (48 x 300 x 300) \\ \hline
        app vector x & Param (48 x 300 x 1) & Param (48 x 300 x 1) \\ \hline
        app vector y & Param (48 x 300 x 1) & Param (48 x 300 x 1) \\ \hline
        app vector z & Param (48 x 300 x 1) & Param (48 x 300 x 1) \\ \hline
        basis mat & Linear (144, 12, bias=False) & Linear (144, 27, bias=False) \\ \hline
        last\_layer & Linear (99, 64, bias=True) & ~ \\
          & ReLU (inlace=True) & ~ \\
          & Linear (64, 64, bias=True) & Spherical Harmonics \\
        ~ & ReLU (inlace=True) & ~ \\
        ~ & Linear (64, 3, bias=True) \\
    \bottomrule
    \end{tabular}
\end{table*}

%% file: tables/supp_quality_speed.tex
\setlength{\tabcolsep}{4pt}
\begin{table}
\centering
\small
\caption{Trade-off between rendering speed and quality with different appearance network capacity. 8$\times$ MS: 8$\times$ sample per-pixel MSAA, 16$\times$ SS: 16$\times$ sample per-pixel SSAA.}
\label{tab:trade-off}
\vspace{-1em}
\begin{tabular}{l|c|ccc|c}
\toprule
Params & AA & PSNR$\uparrow$ & SSIM$\uparrow$ & LPIPS$\downarrow$ & FPS \\
\hline
\#feat=48 & 8$\times$ MS & 30.34 & 0.949 & 0.062 & 93 \\
mlp=3$\times$64 & 16$\times$ SS & 31.16 & 0.954 & 0.057 & 26 \\
\hline
\#feat=48 & 8$\times$ MS & 29.73 & 0.942 & 0.071 & 322 \\
 mlp=3$\times$16 & 16$\times$ SS & 30.49 & 0.947 & 0.064 & 86 \\
\hline
\#feat=12 & 8$\times$ MS & 30.11 & 0.946 & 0.066 & 98 \\
mlp=3$\times$64 & 16$\times$ SS & 30.90 & 0.951 & 0.060 & 27 \\
\hline
\#feat=12 & 8$\times$ MS & 29.55 & 0.941 & 0.073 & 585 \\
 mlp=3$\times$16 & 16$\times$ SS & 30.28 & 0.946 & 0.066 & 163 \\
\hline
\#feat=48 & 8$\times$ MS & 29.73 & 0.943 & 0.068 & 312 \\
 SH & 16$\times$ SS & 30.44 & 0.949 & 0.063 & 82 \\
 \bottomrule
\end{tabular}

\vspace{-1em}
\end{table} 

%% file: tables/supp_app.tex
\begin{table}
\centering
\small
\caption{Ablation study for Stage 2. Except for the first row using GT mesh, the rest experiments are conducted on fixed meshes extracted from pre-trained TensoRF by Marching Cubes. MLP: vanilla NeRF~\cite{mildenhall2021nerf} representation; Hash: HashGrid used in iNGP~\cite{muller2022instant}; SH: Spherical Harmonics; TF: TensoRF-VM.}
    \label{tab:supp_app}
\vspace{-1em}
\begin{tabular}[t]{l|ccc}
\toprule
      Geo. + App. & PSNR$\uparrow$ & SSIM$\uparrow$ & LPIPS$\downarrow$ \\
      \hline
      GT + TF & 31.78 & 0.958 & 0.053 \\
      \hline
      TFmesh + MLP & 26.28 & 0.915 & 0.203 \\
      TFmesh + Hash & 26.62 & 0.921 & 0.090 \\
      TFmesh + SH & 26.48 & 0.909 & 0.103 \\
      TFmesh + TF & 27.00 & 0.929 & 0.081 \\
\bottomrule
    \end{tabular}
\vspace{-1em}
\end{table} 